\documentclass[sigconf]{acmart}

\AtBeginDocument{%
  \providecommand\BibTeX{{%
    \normalfont B\kern-0.5em{\scshape i\kern-0.25em b}\kern-0.8em\TeX}}}

\setcopyright{acmcopyright}
\copyrightyear{2021}
\setcopyright{iw3c2w3}
\acmYear{2021}
\acmConference[WWW '21]{Proceedings of the Web Conference 2021}{April 19--23, 2021}{Ljubljana, Slovenia}
\acmBooktitle{Proceedings of the Web Conference 2021 (WWW '21), April 19--23, 2021, Ljubljana, Slovenia}
\acmPrice{}
\acmDOI{10.1145/3442381.3449934}
\acmISBN{978-1-4503-8312-7/21/04}

\usepackage{subfigure}
\usepackage{xspace}
\usepackage{multirow}
\usepackage{xcolor}
\usepackage{amsmath}
\usepackage{amsfonts}
\usepackage{bm}
\usepackage{soul}
\usepackage{algorithm}
\usepackage{algpseudocode}
\usepackage{booktabs}
\usepackage[normalem]{ulem}        
\usepackage[switch]{lineno} 

\newcommand{\hao}[1]{{\color{black}{#1}}}
\newcommand{\liu}[1]{{\color{black}{#1}}}
\newcommand{\jia}[1]{{\color{black}{#1}}}

\newcommand{\eat}[1]{}

\newcommand{\beijing}{{\sc Beijing}\xspace}
\newcommand{\shanghai}{{\sc Shanghai}\xspace}

\newcommand{\stddpg}{{\sc Master}\xspace}

\newcommand{\eg}{\emph{e.g.},\xspace}
\newcommand{\ie}{\emph{i.e.},\xspace}

\newcommand\figref[1]{Figure~\ref{#1}}

\newcommand\tabref[1]{Table~\ref{#1}}
\newcommand\secref[1]{Section~\ref{#1}}
\newcommand\appref[1]{Appendix~\ref{#1}}
\newcommand\equref[1]{Eq.~(\ref{#1})}

\newtheorem{pro}{Problem}
\newtheorem{defi}{Definition}

\begin{document}
\settopmatter{printacmref=true}

\title{Intelligent Electric Vehicle Charging Recommendation\\ Based on Multi-Agent Reinforcement Learning}

\author{Weijia Zhang$^{1\dagger}$, Hao Liu$^{2*}$, Fan Wang$^{3}$, Tong Xu$^{1}$, Haoran Xin$^{1}$, Dejing Dou$^{2}$, Hui Xiong$^{4*}$}

\affiliation{
	$^1$ University of Science and Technology of China,
	$^2$ Business Intelligence Lab, Baidu Research, \\
	$^3$ Baidu Inc.,
	$^4$ Rutgers University\\
	\{wjzhang3,xinhaoran\}@mail.ustc.edu.com,
	\{liuhao30,wangfan04,doudejing\}@baidu.com,\\
	tongxu@ustc.edu.com,
	hxiong@rutgers.edu
}
\thanks{
    $^{*}$~Corresponding author.\\
    $^{\dagger}$~The research was done when the first author was an intern in Baidu Research under the supervision of the second author.}


\begin{abstract}
	Electric Vehicle~(EV) has become a preferable choice in the modern transportation system due to its environmental and energy sustainability.
	However, in many large cities, EV drivers often fail to find the proper spots for charging, because of the limited charging infrastructures and the spatiotemporally unbalanced charging demands.
	Indeed, the recent emergence of deep reinforcement learning provides great potential to improve the charging experience from various aspects over a long-term horizon.
	In this paper, we propose a framework, named \emph{\underline{M}ulti-\underline{A}gent \underline{S}patio-\underline{Te}mporal \underline{R}einforcement Learning}~(\stddpg), for intelligently recommending public accessible charging stations by jointly considering various long-term spatiotemporal factors.
	Specifically, by regarding each charging station as an individual agent, we formulate this problem as a multi-objective multi-agent reinforcement learning task.
	We first develop a multi-agent actor-critic framework with the centralized attentive critic to coordinate the recommendation between geo-distributed agents.
	Moreover, to quantify the influence of future potential charging competition, we introduce a delayed access strategy to \hao{exploit the knowledge of future charging competition during training}.
	After that, to effectively optimize multiple learning objectives, \hao{we extend the centralized attentive critic to multi-critics and develop a} dynamic gradient re-weighting strategy to adaptively guide the optimization direction.	
	Finally, extensive experiments on two real-world datasets demonstrate that \stddpg achieves the best comprehensive performance compared with \hao{nine} baseline approaches.
\end{abstract}



\maketitle

\section{Introduction}\label{sec:intro}

Due to the low-carbon emission and energy efficiency, Electric vehicles (EVs) are emerging as a favorable choice in the modern transportation system to meet the increasing environmental concerns~\liu{\cite{han2020joint}} and energy insufficiency~\cite{savari2020internet}.
In 2018, there are over 2.61 million EVs on the road in China, and this number will reach 80 million in 2030~\cite{zheng2020systematic}.
Although the government is expanding the publicly accessible charging network to fulfill the explosively growing on-demand charging requirement, it is still difficult for EV drivers to charge their vehicles due to the fully-occupied stations and long waiting time~\cite{wang2020faircharge,tian2016real,li2015growing,du2018demand}.
Undoubtedly, such an unsatisfactory charging experience raises \hao{undesirable charging cost and inefficiency}, and may even increase the EV driver's "range anxiety", which prevents the further prevalence of EVs.
Therefore, it is appealing to provide intelligent charging recommendations to improve the EV drivers' charging experience from various aspects, such as minimize the charging wait time~(CWT), reduce the charging \jia{price~(CP)}, as well as optimize the charging failure rate~(CFR) to improve the efficiency of the global charging network.

The charging recommendation problem distinguishes itself from traditional recommendation tasks\liu{~\cite{guo2020survey,xin2021out}} from two perspectives.
First, the number of charging spots in a target geographical area may be limited, which may induce potential \hao{resource competition} between EVs.
\hao{Second, depending on the battery capacity and charging power, the battery recharging process may block the charging spot for several hours}.
As a result, the current recommendation may influence future EV charging recommendations and produce a long-term effect on the global charging network.
\hao{Previously, some efforts \cite{wang2020faircharge,guo2017recommendation,tian2016real,wang2019tcharge,cao2019toward,yang2014optimal} have been made for charging recommendations via greedy-based strategies by suggesting the most proper station for the current EV driver in each step concerning a single objective~(\eg minimizing the overall CWT).
However, such an approach overlooks the long-term contradiction between the space-constrained charging capacity and the spatiotemporally unbalanced charging demands, which leads to sub-optimal recommendations from a global perspective~(\eg longer overall CWT and higher CFR).}

Recently, Reinforcement Learning~(RL) has shown great advantages in optimizing sequential decision problems in the dynamic environment, such as order dispatching for ride-hailing and shared-bike rebalancing~\cite{li2019efficient,li2018dynamic}. 
By interacting with the environment, the agent in RL learns the policy to achieve the global optimal long-term reward\liu{~\cite{fan2020autofs}}.
Therefore, it is intuitive for us to improve charging recommendations based on RL, with long-term goals such as minimizing the overall CWT, the average \jia{CP}, and the CFR.
However, there exist several technical challenges in achieving this goal.

The first challenge comes from the \emph{large state and action space}.
There are millions of EVs and thousands of publicly accessible charging stations in a metropolis such as Beijing. 
Directly learning a centralized agent system across the city requires handling large state and action space and high-dimensional environment, which will induce severe scalability and efficiency issues~\cite{papoudakis2019dealing,chu2019multi}.
Moreover, the "single point of failure"~\cite{lynch2009single} in the centralized approach may fail the whole system~\cite{li2019efficient}.
As an alternative, an existing approach~\citet{wang2020joint} tried to model a small set of vehicles as multiple agents and maximize the cumulative reward in terms of the number of served orders. 
However, in our charging recommendation task, most charging requests are ad-hoc and from non-repetitive drivers, which renders it impossible to learn a dedicated policy for each individual EV.
To address the first challenge, we regard each charging station as an individual agent and formulate EV charging recommendation as a multi-agent reinforcement learning~(MARL) task, and we propose a multi-agent actor-critic framework. In this way, each individual agent has a quantity-independent state action space, which can be scaled to more complicated environments and is more robust to other agents' potential failure.

The second challenge is the \emph{coordination and cooperation in the large-scale agent system}.
For a charging request, only one station will \hao{finally serve for charging}.
Different agents should be coordinated to \hao{achieve} better recommendations.
Moreover, the cooperation between agents is the key for long-term recommendation optimization.
For example, consider a heavily occupied charging station with a large number of incoming charging requests. Other stations with sufficient available charging spots can help balance the charging demands via cross-agent cooperation.
\hao{To tackle this challenge, we \jia{analogy} the process of agents \jia{taking} actions to a bidding game~\cite{zhao2018deep}, and propose a tailor designed centralized attentive critic module to stimulate multiple agents to learn globally coordinated and cooperative policies.}

\begin{figure}[tb]
	\centering
	\includegraphics[width=0.95\columnwidth]{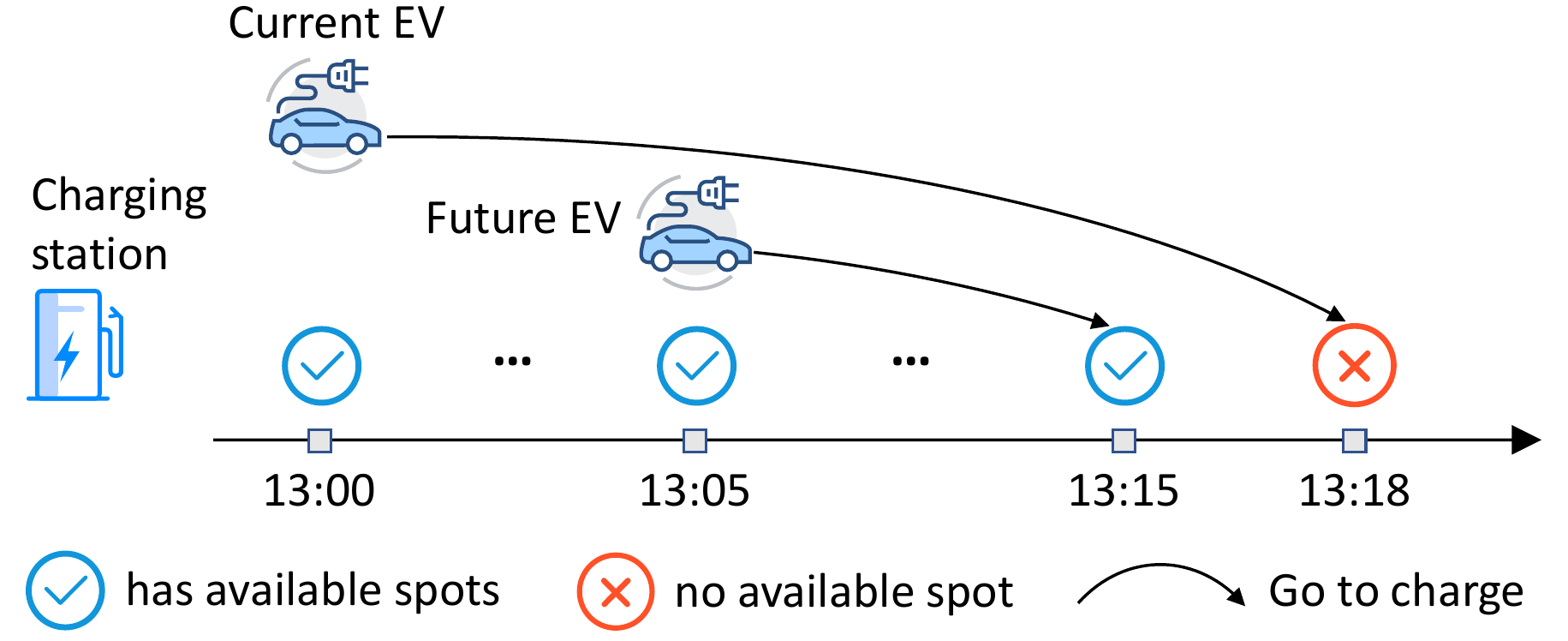}
	\caption{Illustrative example of future potential charging competition between EVs. The last available charging spot is preoccupied by a future EV, leading to extra CWT for current EV.}
	\label{fig:tmp_example}
\end{figure}
The third challenge is the \emph{potential competition of future charging requests}.
As illustrated in \figref{fig:tmp_example}, the competition comes from the temporally distributed charging requests for the limited charging resource.
In the real world, the potential competition of future charging requests may happen in arbitrary charging stations, leading to problems such as extra CWT and charging failure.
However, it is hard to quantify the influence of such future charging requests in advance.
\hao{To this end, we integrate the future charging competition information into centralized attentive critic module through a delayed access strategy and transform our framework to a centralized training with decentralized execution architecture to enable online recommendation. In this way, the agents can fully harness future knowledge in training phase and take actions immediately without requiring future information during execution.}


Finally, it is challenging to \emph{jointly optimize multiple optimization objectives}.
As aforementioned, it is desirable to simultaneously consider various long-term goals, such as the overall CWT, average \jia{CP}, and the CFR.
\hao{However, these objectives may in different scales and lie on different manifold structures.
Seamlessly optimizing multiple objectives may lead to poor convergence and induce sub-optimal recommendations for certain objectives.
Therefore, we extend the centralized attentive critic module to multi-critics for multiple objectives and develop a dynamic gradient re-weighting strategy to adaptively guide the optimization direction by forcing the agents to pay more attention to these poorly optimized objectives.}

Along these lines, in this paper, we propose the \emph{\underline{M}ulti-\underline{A}gent \underline{S}patio-\underline{Te}mporal \underline{R}einforcement Learning}~(\stddpg) framework for intelligent charging recommendation. Our major contributions are summarized as follows:
\hao{(1) We formulate the EV charging recommendation problem as a MARL task. To the best of our knowledge, this is the first attempt to apply MARL for multi-objective intelligent charging station recommendations.
(2) We develop the multi-agent actor-critic framework with centralized training decentralized execution. In particular, the centralized attentive critic achieves coordination and cooperation between multiple agents globally and exploits future charging competition information \jia{through a delayed access strategy} during model training. Simultaneously, the decentralized execution of multiple agents avoids the large state action space problem and enables immediate online recommendations without requiring future competition information.
(3) We extend the centralized attentive critic to multi-critics to support multiple optimization objectives. By adaptively guiding the optimization direction via a dynamic gradient re-weighting strategy, we improve the learned policy from potential local optima induced by dominating objectives.
(4) We conduct extensive experiments on two real-world datasets collected from \beijing and \shanghai. The results demonstrate our model achieves the best comprehensive performance against nine baselines.}

\section{Problem definition}\label{sec:preliminary}
In this section, we introduce some important definitions and formalize the EV charging recommendation problem. 

Consider a set of $N$ charging stations $C=\{c^1,c^2,\dots,c^N\}$, by regarding each day as an episode, we first define the charging request as below.
\begin{defi}
	\textbf{Charging request}. A charging request $q_t =\langle l_t, \\T_t, T^c_t \rangle \in Q$ is defined as the $t$-th request~(\ie step $t$) of a day. 
	Specifically, $l_t$ is the location of $q_t$, $T_t$ is the real-world time of the step $t$, and $T_t^c$ is the real-world time $q_t$ finishes the charging request. 
\end{defi}
We say a charging request is finished if it successfully takes the charging action or finally gives up~(\ie charging failure). We further denote $|Q|$ as the cardinality of $Q$.
In the following, we interchangeably use $q_t$ to denote the corresponding EV of $q_t$.

\begin{defi}
	\textbf{Charging wait time~(CWT)}. The charging wait time is defined as the summation of travel time from the location $l_t$ of charging request $q_t$ to the target charging station $c^i$ and the queuing time at $c^i$ until $q_t$ is finished.
\end{defi}
\begin{defi}
	\textbf{Charging price~(CP)}. The charging price is defined as the unit price per kilowatt hour~(kWh). In general, the charging price is a combination of electricity cost and service fee.
\end{defi}
\begin{defi}
	\textbf{Charging failure rate~(CFR)}. The charging failure rate is defined as the ratio of
	the number of charging requests who accept our recommendation but fail to charge over the total number of charging requests who accept our recommendation.
\end{defi}
\begin{pro}
	\textbf{EV Charging Recommendation}.
	Consider a set of charging requests $Q$ in a day, our problem is to recommend each $q_t \in Q$ to the most proper charging station $rc_t \in C$, with the long-term goals of simultaneously minimizing the overall CWT, average CP, and the CFR for the $q_t \in Q$ who accept our recommendation.
\end{pro}

\section{Methodology}
\hao{In this section, we present the MARL formulation for the EV charging recommendation task and detail our \stddpg framework with centralized training decentralized execution~(CTDE). Moreover, we elaborate on the generalized multi-critic architecture for multiple objectives optimization.}


\subsection{MARL Formulation}\label{sec:formulation}
\begin{figure}[tb]
	\centering
	\hspace{2mm}
	\includegraphics[width=0.95\columnwidth]{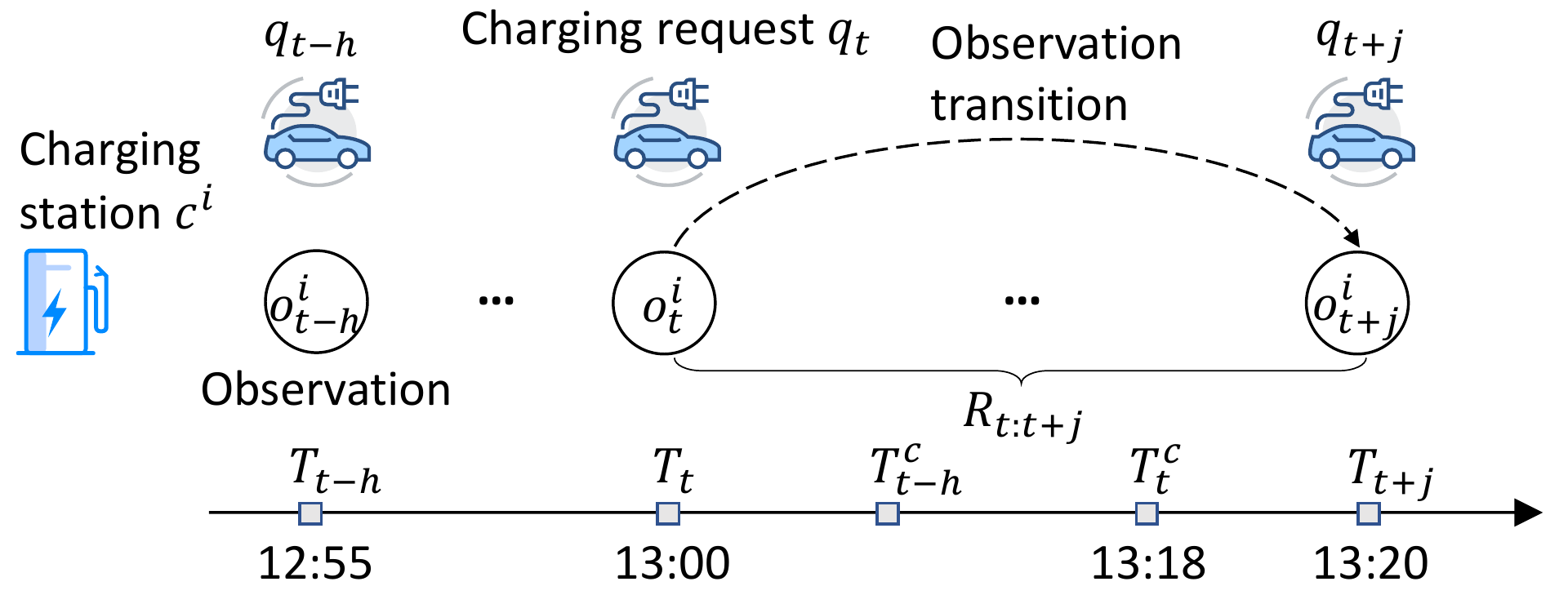}
	\caption{Illustrative  example  of transition in \stddpg.}
	\label{fig:trans}
\end{figure}

We first present our MARL formulation for the EV Charging Recommendation task. 
\begin{itemize}
	\item \textbf{Agent} $c^i$. In this paper, we regard each charging station $c^i \in C$ as an individual agent.
	Each agent will make timely recommendation decisions for a sequence of charging requests $Q$ that keep coming throughout a day with multiple long-term optimization goals.
	
	\item \textbf{Observation $o_t^i$}. 
	Given a charging request $q_t$, we define the observation $o_t^i$ of agent $c^i$ as a combination of the index of $c^i$, the real-world time $T_t$, the number of current available charging spots of $c^i$~(supply), the number of charging requests around $c^i$ in the near future~(future demand), the charging power of $c^i$, the estimated time of arrival~(ETA) from location $l_t$ to $c^i$, and the CP of $c^i$ at the next ETA. We further define $s_t=\{o_t^1,o_t^2,\dots,o_t^N\}$ as the state of all agents at step $t$.
	
	\item \textbf{Action $a_t^i$}. 
	Given an observation $o_t^i$, an intuitional design for the action of agent $c^i$ is a binary decision,~\ie~recommending $q_t$ to itself for charging or not. However, because one $q_t$ can only choose one station for charging, \hao{multiple agents' actions may be tied together and are difficult to coordinate}.
	Inspired by the bidding mechanism~\cite{zhao2018deep}, \hao{we design each agent $c^i$ offers a scalar value to "bid" for $q_t$ as its action $a_t^i$.} 
	By defining $u_t=\{a_t^1,a_t^2,\dots,a_t^N\}$ as the joint action, $q_t$ will be recommended to the agent with the highest "bid" value, \ie $rc_t=c^i$, where $i=\arg\max(u_t)$.
	
	\item \textbf{Transition}. 
	For each agent $c^i$, its observation transition is defined as the transition from the current charging request $q_t$ to the next charging request $q_{t+j}$ after $q_t$ is finished.
	Let's elaborate on this via an illustrative example as shown in \figref{fig:trans}.
	Consider a charging request $q_t$ arises at  $T_t$~(13:00). 
	At this moment, each agent $c^i$ takes action $a_t^i$ based on its observation $o_t^i$ and jointly decide the recommended station $rc_t$.
	After the request finish time $T^c_t$~(13:18), the subsequent charging request $q_{t+j}$ will arise at $T_{t+j}$~(13:20).
	In this case, the observation transition of agent $c^i$ 
	is defined as $(o_t^i,a_t^i,o_{t+j}^i)$, where $o_t^i$ is the current observation, $o_{t+j}^i$ is the observation corresponding to $q_{t+j}$.
	
	\item \textbf{Reward}. 
	In our MARL formulation, we propose a lazy reward settlement scheme~(\ie return rewards when a charging request is finished), and integrate three goals into two \jia{natural} reward functions. Specifically, if a charging request $q_t$ succeeds in charging, then the environment will return the negative of CWT and negative of CP as the part of reward $r^{cwt}(s_t, u_t)$ and reward $r^{cp}(s_t, u_t)$ respectively.
	For the case that the CWT of $q_t$ exceeds a threshold, the recommendation will be regarded as failure and environment will return much smaller rewards as penalty to stimulate agents reducing the CFR. Overall, we define two immediate rewards function for three goals as
	\begin{equation}
	\label{equ:reward_cwt}
	r^{cwt}(s_t, u_t)=\left\{
	\begin{aligned} 
	&-CWT,\quad charging~success\\
	&\quad\,\,\epsilon_{cwt},\quad charging~failure
	\end{aligned}~,
	\right.
	\end{equation}
	\begin{equation}
	\label{equ:reward_cp}
	r^{cp}(s_t, u_t)=\left\{
	\begin{aligned} 
	&-CP,\quad charging~success\\
	&\quad \epsilon_{cp},\quad charging~failure
	\end{aligned}~,
	\right.
	\end{equation}
	where $\epsilon_{cwt}$ and $\epsilon_{cp}$ are penalty rewards. 
	\hao{All agents in our model share the unified rewards, \jia{means agents make the recommendation decisions cooperatively.}
	Since the observation transition from $o_t^i$ to $o_{t+j}^i$ may cross multiple lazy rewards~(\eg $T_{t-h}^c$ and $T_{t}^c$ as illustrated in \figref{fig:trans}), we calculate the cumulative discounted reward by summing the rewards of all recommended charging requests $q_{t'}$}~(\eg $q_{t-h}$ and $q_{t}$) whose $T_{t'}^c$ is between $T_t$ and $T_{t+j}$, denoted by 
	\begin{equation}\label{equ:acc_reward}
	R_{t:t+j}=\sum_{T_t<T_{t'}^c\leq T_{t+j}} \gamma^{(T_{t'}^c-T_t-1)}r(s_{t'},u_{t'}),
	\end{equation}
	where $\gamma$ is the discount factor, 
	\hao{and $r(\cdot,\cdot)$ can be one of the two reward functions or the average of them depending on the learning objectives.}
\end{itemize}

\subsection{Centralized Training Decentralized Execution}
Centralized training decentralized execution is a class of methods in MARL to stimulate agents to learn coordinated policies and address the non-stationary environment problem~\cite{lowe2017multi,DBLP:conf/aaai/FoersterFANW18,iqbal2019actor}. 
\hao{In \stddpg, the CTDE consists of three modules, the centralized attentive critic, the delayed access information strategy to integrate future charging competition, and the decentralized execution process.
The advantages of CTDE for EV charging recommendation are two folds.
On one hand, the centralized training process can motivate multiple agents to learn cooperation and other specific policies by perceiving a more comprehensive landscape and exploiting future information in hindsight. 
On the other hand, the execution process is fully decentralized without requiring complete information in the training phase, which guarantees efficiency and flexibility in online recommendation. 
}

\subsubsection{\textbf{Centralized Attentive Critic}}\label{sec:centralized}
\begin{figure}[tb]
	\centering
	\hspace{-2mm}
	\includegraphics[width=1\columnwidth]{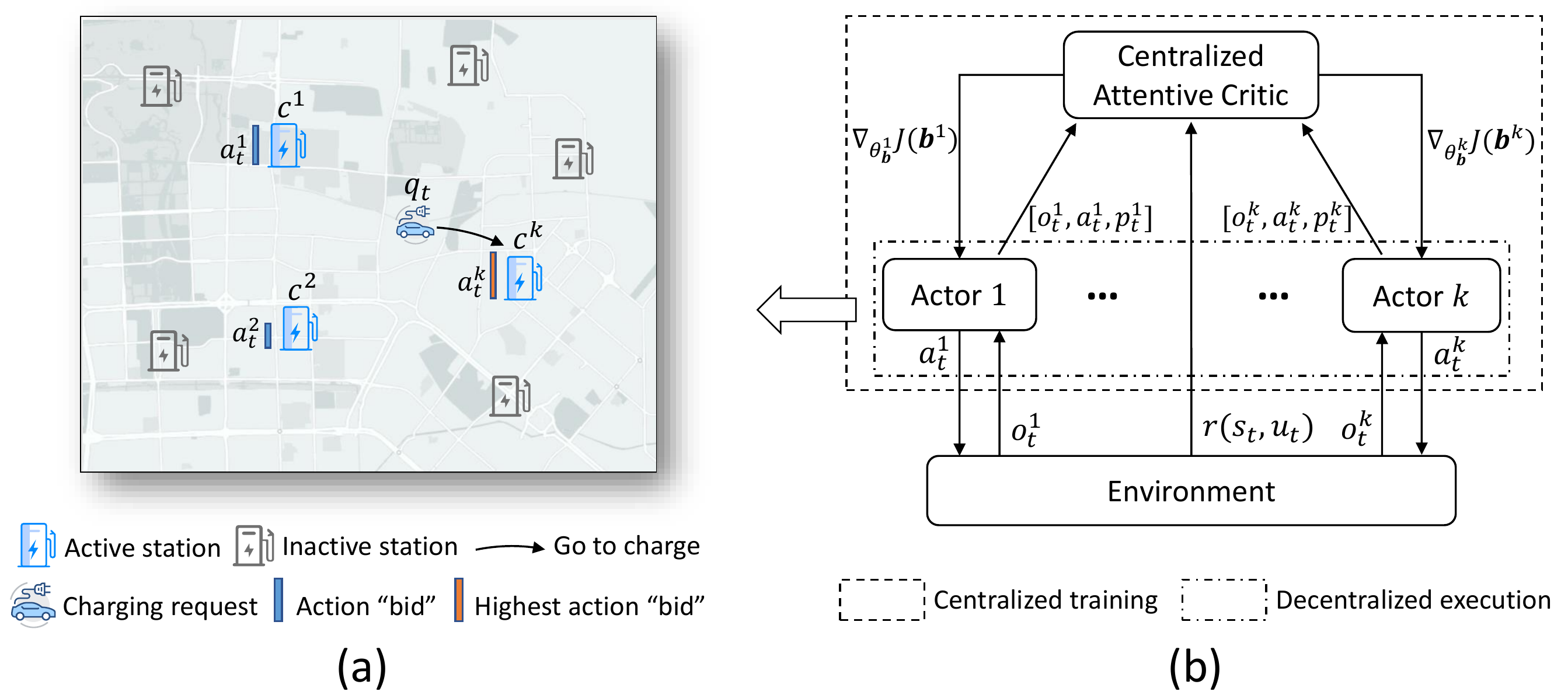}
	\caption{(a)~Decentralized execution of active agents. The charging request will be recommended to the active charging station with the highest action "bid". (b)~The centralized training decentralized execution process of \stddpg. }
	\label{fig:overall}
\end{figure}

\hao{To motivate the agents to make recommendations cooperatively, we devise the multi-agent actor-critic framework with a centralized attentive critic for deterministic policies learning.} 
A similar MARL algorithm with CTDE architecture is proposed in~\cite{lowe2017multi}, which incorporates the full state $s_t$ and joint action $u_t$ of all agents into the critic to motivate agents to learn coordinated and cooperative policies. However, such an approach suffers from the large state and action space problem in our task.

\hao{In practice, the EVs tend to go to nearby stations for charging. Based on this fact, given a charging request $q_t$, we only activate the agents nearby $q_t$~(\eg top-$k$ nearest $q_t$) to take actions, denoted as $C_t^a$.
We set other agents who are far away from $q_t$ inactive and don't participate in the recommendation for $q_t$, as illustrated in \figref{fig:overall}(a). 
In this way, only a small number of active agents are involved to learn cooperation for better recommendations.
However, one intermediate problem is that the active agents of different $q_t$ are usually different. To this end, we propose to use the attention mechanism \jia{which is permutation-invariant to integrate information of the active agents}.}
\hao{Specifically, the attention mechanism automatically quantifies the influence of each active agents by}
\begin{equation}\label{equ:att_weight}
e_t^i = \mathbf{v}^{\top}\text{tanh}\left(\mathbf{W}_a\left[o_{t}^i \oplus a_{t}^i \oplus p_t^i\right]\right),
\end{equation}
\begin{equation}
\alpha_{t}^i = \frac{exp(e_{t}^i)}{\sum_{j \in C_t^a}exp(e_{t}^j)},
\end{equation}
where $\mathbf{v}$ and $\mathbf{W}_a$ are learnable parameters, $\oplus$ is the concatenation operation, and $p_{t}^i$ are \hao{future} information that will be detailed in \secref{sec:competition}.
Once the influence weight $\alpha_t^i$ of each active agent $c^i \in C_t^a$ is obtained, we can derive the attentive representation \hao{of all active agents by}
\begin{equation}\label{equ:att_sum}
x_{t} = \text{ReLU}\left(\mathbf{W}_{c}\sum_{i \in C_t^a} \alpha_{t}^i \left[o_{t}^i \oplus a_{t}^i \oplus p_t^i\right]\right),
\end{equation}
where $\mathbf{W}_c$ are learnable parameters.

\hao{Given the state $s^a_t$, joint action $u^a_t$, and future information $p_t$ of active agents,}
the actor policy of each agent $c^i \in C^a_t$ can be updated by the gradient of the expected return following the chain rule, which can be written by
\begin{equation}\label{equ:gradient}
\nabla_{\theta^i_{\bm{b}}}J(\bm{b}^i)=\mathbb{E}_{s^a_t,p_t \sim D}\left[\nabla_{\theta^i_{\bm{b}}}\bm{b}^i(a^i_t|o^i_t)\nabla_{a_t^i} Q_{\bm{b}}(x_t)|_{a_t^i=\bm{b}^i(o_t^i)}\right],
\end{equation}
where $\theta^i_{\bm{b}}$ are learnable parameters of actor policy $\bm{b}^i$ of agent $c^i$, $D$ is the experience replay buffer containing the transition tuples $(s_t^a,u_t^a,p_t,s_{t+j}^a,p_{t+j},R_{t:t+j})$.
Then, each agent updates its policy by the gradients propagated back from the centralized attentive critic. The entire process is shown in \figref{fig:overall}(b).
\jia{As the centralized attentive critic perceived more complete information of all active agents, it can motivate the agents to learn policies in a coordinated and cooperative way.}
The centralized attentive critic $Q_{\bm{b}}$ is updated by minimizing the following loss:
\begin{equation}
\label{equ:gradient1}
L(\theta_Q)=\mathbb{E}_{s_t^a,u_t^a,p_t,s_{t+j}^a,p_{t+j},R_{t:t+j} \sim D}\left[\left(Q_{\bm{b}}(x_t)-y_t\right)^2\right],\\
\end{equation}
\begin{equation}
\label{equ:gradient2}
y_t=R_{t:t+j}+\gamma^{(T_{t+j}-T_{t})}Q^{'}_{\bm{b}^{'}}(x_{t+j})|_{a_{t+j}^i=\bm{b}^{'i}(o_{t+j}^i)}, 
\end{equation}
where $\theta_Q$ are the learnable parameters of critic $Q_{\bm{b}}$, $\bm{b}^{'i}$ and $Q^{'}_{\bm{b}^{'}}$ are the target actor policy of $c^i$ and target critic function with delayed parameters $\theta_{\bm{b}}^{'i}$ and $\theta_Q^{'}$.

\subsubsection{\textbf{Integrate Future Charging Competition}}\label{sec:competition}
The public accessible charging stations are usually first-come-first-served, which induces future potential charging competition between the \hao{continuously arriving} EVs. 
\hao{Recommending charging requests without considering future potential competition will lead to extra CWT or even charging failure.
However, it is a non-trivial task to incorporate the future competition since we can not know the precise amount of forthcoming EVs and available charging spots at a future step in advance.
} 
\eat{due to the future uncertainties.} 

\hao{In this work, we further extend the centralized attentive critic with a delayed access strategy to harness the future charging competition information, so as to enable the agents learning policies with future knowledge in hindsight.}
\hao{Specifically, consider a charging request $q_t$, without changing the execution process, we postpone the access of transition tuples until the future charging competition information in transition with respect to $q_t$ is fully available.}
\hao{Concretely, we} extract the accurate number of available charging spots of $c^i$ at every next $d$ minutes after $T_t$ to reflect the future potential charging competition for $q_t$, denoted as $I^i_t$. 
\hao{
Note that \jia{we erase the influence of $q_t$ for $I^i_t$,} and the number of available charging spots can be negative, means the number of EVs queuing at the station.}
\eat{For simplicity, we omit the potential influence of $q_t$ to $I^i_t$, which we empirically find that the performance is insensitive to it.} 
Overall, we obtain the future charging competition information of each $c^i$ for $q_t$ via a fully-connected layer
\begin{equation}\label{equ:competition}
p_t^i = \text{ReLU}\left(\mathbf{W}_p I_t^i\right),
\end{equation}
where $\mathbf{W}_p$ are the learnable parameters. The $p_t^i$ is integrated into the centralized attentive critic~(\equref{equ:att_weight}$\sim$\equref{equ:att_sum}) as the \jia{enhanced} information to facilitate the \hao{agents' cooperative policy learning}.

\subsubsection{\textbf{Decentralized Execution}}\label{sec:execution}
The execution process is fully decentralized, by only invoking the learned actor policy with its own observation.
Specifically, for a charging request $q_t$, the agent $c^i\in C^a_t$ takes action $a^i_t$ based on its $o^i_t$ by
\begin{equation}\label{equ:competition}
a^i_t = \bm{b}^i(o_t^i).
\end{equation}
And $q_t$ will be recommended to the active agent with the highest $a^i_t$ among all the actions of $C^a_t$.
The execution of each agent is lightweight and does not require future charging competition information.
More importantly, the large-scale agent system is fault-tolerant even part of the agents are failing.


\subsection{Multiple Objectives Optimization}\label{sec:objectives}
\begin{figure}[t]
\centering
\subfigure[{\small Reward about CWT}]{
	\includegraphics[width=0.493\columnwidth]{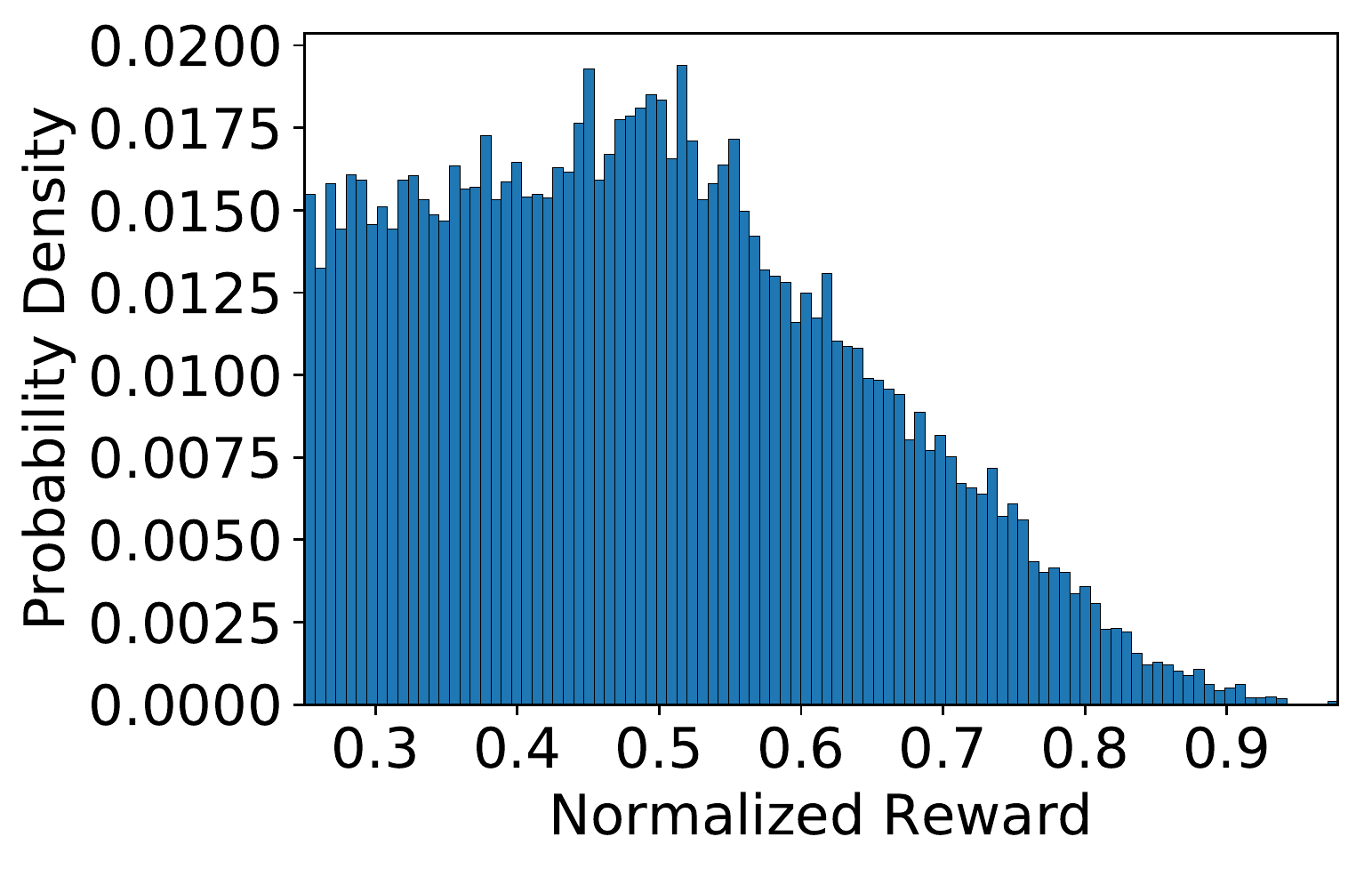}}
\subfigure[{\small Reward about CP}]{
    \label{fig:cpreward_dist}
	\includegraphics[width=0.47\columnwidth]{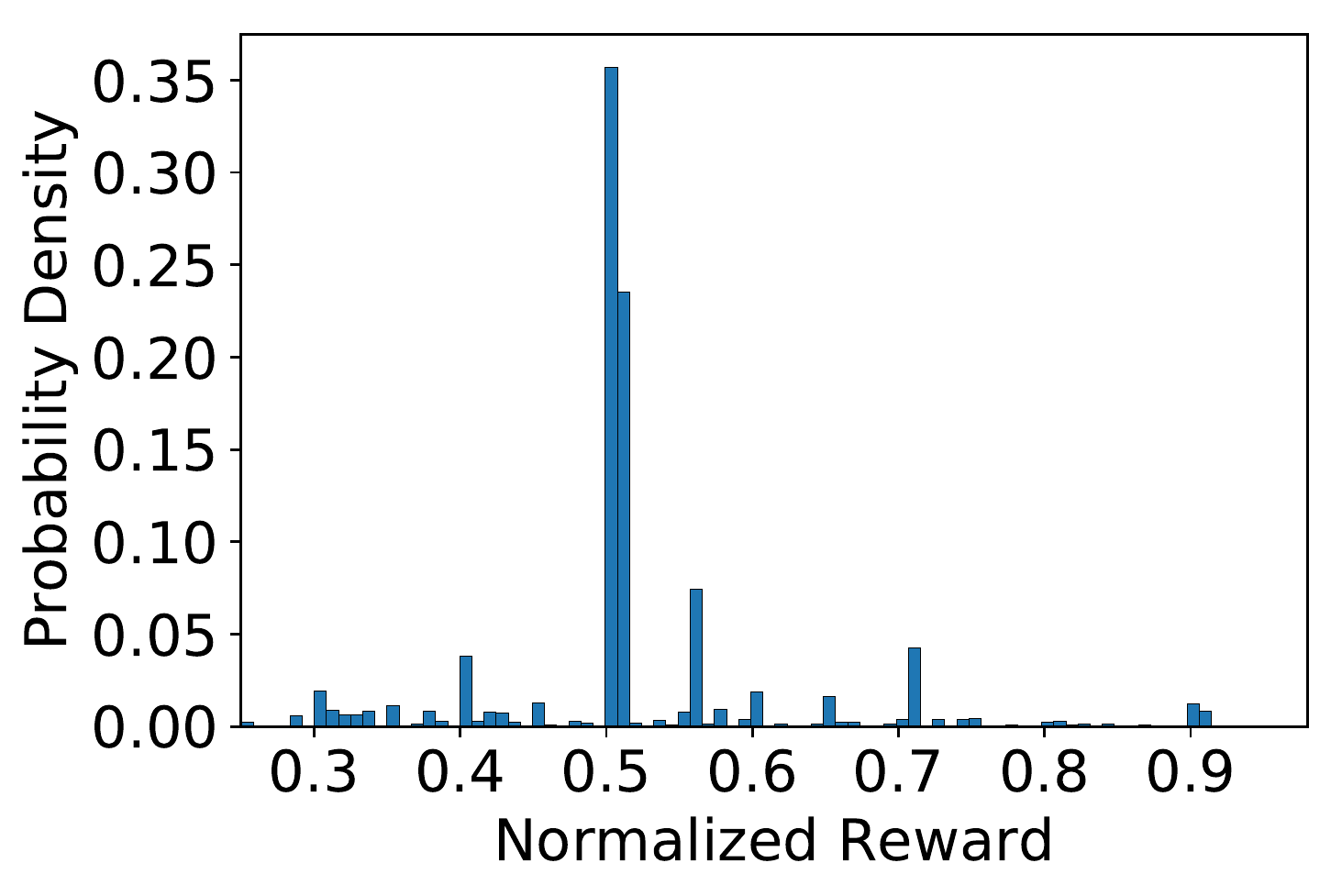}}
\caption{Distributions of two normalized rewards.}
\label{fig:reward_dist}
\end{figure} 

\begin{algorithm}[tb] 
	\caption{ \stddpg algorithm} 
	\label{alg:master} 
	\begin{algorithmic}[1] 
		\State Randomly initialize critic networks $Q_{\bm{b}}^{cwt}$, $Q_{\bm{b}}^{cp}$ and  each actor network $\bm{b}^i$ with weights $\theta_Q^{cwt}$, $\theta_Q^{cp}$ and $\theta_{\bm{b}}^i$.
		\State Initialize target networks $Q_{\bm{b}^{'}}^{cwt'}$, $Q_{\bm{b}^{'}}^{cp'}$ and $\bm{b}^{'i}$ with weights $\theta_Q^{cwt'} \leftarrow \theta_Q^{cwt}$, $\theta_Q^{cp'} \leftarrow \theta_Q^{cp}$, $\theta_{\bm{b}}^{'i} \leftarrow \theta_{\bm{b}}^i$.
		\State Initialize objective-specific optimal networks $Q_{\bm{b}_{cwt}^{^*}}^{cwt^*}$, $Q_{\bm{b}_{cp}^{^*}}^{cp^*}$, $\bm{b}_{cwt}^{^*i}$ and $\bm{b}_{cp}^{^*i}$ with well-trained weights $\theta_Q^{cwt^*}$, $\theta_Q^{cp^*}$, $\theta_{\bm{b}_{cwt}}^{^*i}$ and $\theta_{\bm{b}_{cp}}^{^*i}$.
		\State Initialize replay buffer $D$.
		\For {$m=1$ to max-iterations}
		\State Reset environment.
		
		\For {$t^{-}=1$ to $|Q|$}
			\For {agent $c^i\in C^a_{t^{-}}$}
			\State Take action $a_{t^{-}}^i = \bm{b}^i(o^i_{t^{-}})$ for charging request $q_{t^{-}}$.
			\EndFor
		\State Store available delayed transition tuples $(s_{t'}^a,u_{t'}^a,p_{t'},$
		\State  $s_{t'+j'}^a,p_{t'+j'},R_{t':t'+j'}^{cwt},R_{t':t'+j'}^{cp})$ into $D$. 
		\State Sample a random minibatch of $M$ transitions
		\State  $(s_{t}^a,u_{t}^a,p_{t},s_{t+j}^a,p_{t+j},R_{t:t+j}^{cwt},R_{t:t+j}^{cp})$ from $D$.
		\State Set $\left.y_t^{cwt}=R_{t:t+j}^{cwt}+\gamma^{(T_{t+j}-T_{t})}Q^{cwt'}_{\bm{b}^{'}}(x_{t+j})|_{a_{t+j}^i=\bm{b}^{'i}(o_{t+j}^i)}\right.$.
		\State Set $y_{t}^{cp}=R_{t:t+j}^{cp}+\gamma^{(T_{t+j}-T_{t})}Q^{cp'}_{\bm{b}^{'}}(x_{t+j})|_{a_{t+j}^i=\bm{b}^{'i}(o_{t+j}^i)}$.
		\State Update Critic $Q_{\bm{b}}^{cwt}$ and $Q_{\bm{b}}^{cp}$ by minimizing the losses: 
		\State \hskip1.5em $L(\theta_Q^{cwt})=\frac{1}{M}\sum_{t}\left(Q_{\bm{b}}^{cwt}(x_{t})-y_{t}^{cwt}\right)^2$.
		\State \hskip1.5em $L(\theta_Q^{cp})=\frac{1}{M}\sum_{t}\left(Q_{\bm{b}}^{cp}(x_{t})-y_{t}^{cp}\right)^2$.
		\State Compute $\beta_t$ through \equref{equ:gap_ratio} and \equref{equ:update_weight}.
		\For {agent $c^i\in C^a_t$}
			\State Update actor by the sampled policy gradient: 
			\State \hskip1.5em $\nabla_{\theta^i_{\bm{b}}}J(\bm{b}^i) \approx \frac{1}{M} \sum_{t} \left(\beta_t\nabla_{\theta^i_{\bm{b}}}\bm{b}^i(a^i_t|o^i_t)\nabla_{a_t^i} Q_{\bm{b}}^{cwt}(x_t)\right.$
			\State \hskip1.5em
			$\left.+(1-\beta_t)\nabla_{\theta^i_{\bm{b}}}\bm{b}^i(a^i_t|o^i_t)\nabla_{a_t^i} Q_{\bm{b}}^{cp}(x_t)\right)|_{a_t^i=\bm{b}^i(o_t^i)}$.
			\State \hskip1.5em $\theta_{\bm{b}}^i \leftarrow \theta_{\bm{b}}^i + \eta \nabla_{\theta^i_{\bm{b}}}J(\bm{b}^i)$.
			\State Update target actor networks: 
			\State \hskip1.5em $\theta_{\bm{b}}^{'i} \leftarrow \tau \theta_{\bm{b}}^i + (1-\tau)\theta_{\bm{b}}^{'i}$.
		\EndFor
		\State Update target critic networks:
		\State \hskip1.5em $\theta_Q^{cwt'} \leftarrow \tau \theta_Q^{cwt} + (1-\tau) \theta_Q^{cwt'}$.
		\State \hskip1.5em $\theta_Q^{cp'} \leftarrow \tau \theta_Q^{cp} + (1-\tau) \theta_Q^{cp'}$.
	\EndFor
	\EndFor
	\end{algorithmic}
\end{algorithm}

The goal of Electric Vehicle Charging Recommendation task is to simultaneously minimize the overall CWT, average CP and the CFR. 
\hao{We integrate these goals into two objectives corresponding to two reward functions, \ie $r^{cwt}$ and $r^{cp}$ as defined in \equref{equ:reward_cwt} and \equref{equ:reward_cp}.
\figref{fig:reward_dist} depicts the normalized reward distributions of $r^{cwt}$ and $r^{cp}$ by a random policy executing on a set of charging requests \jia{that charging success}.
As can be seen, the distribution of different objectives can be very different. 
More importantly, the optimal solution of different objectives may be \jia{divergent}, \eg a cheaper charging station may be very popular and need a longer CWT.
The above facts imply one policy that performs well on one objective may get stuck on \jia{another objective}.
A recommender that is biased to a particular objective is risky to induce unsatisfactory experience to most users.
}
 
\hao{A simple way to optimize multiple objectives is to average the reward of multiple objectives with a set of prior weights and maximize the combined reward as a single objective.
However, such a static approach is inefficient for convergence and may induce a biased solution to a particular objective.
To this end, we develop a dynamic gradient re-weighting strategy to self-adjust the optimization direction to adapt different training stages and enforce the policy to perform well on multiple objectives.
Specifically, we extend the centralized attentive critic to multi-critics, where each critic is corresponding to a particular objective.
Particularly, in our task, we learn two centralized attentive critics, $Q_{\bm{b}}^{cwt}$ and $Q_{\bm{b}}^{cp}$, which correspond to the expected returns of reward $r^{cwt}$ and $r^{cp}$, respectively.}
Since the structure of two critics are identical, we only present $Q_{\bm{b}}^{cwt}$ for explanation, \jia{which is formulated as}
\begin{equation}
\begin{aligned}
Q_{\bm{b}}^{cwt}(x_t)=&\mathbb{E}_{s_{t+j}^a,p_{t+j},R_{t:t+j}^{cwt} \sim E}\left[R_{t:t+j}^{cwt}+\right.\\
&\left. \gamma^{(T_{t+j}-T_{t})}Q^{cwt'}_{\bm{b}^{'}}(x_{t+j})|_{a_{t+j}^i=\bm{b}^{'i}(o_{t+j}^i)} \right],
\end{aligned}
\end{equation}
where $E$ denotes the environment, and $R_{t:t+j}^{cwt}$ is the cumulative discounted reward~(defined in \equref{equ:acc_reward}) with respect to $r^{cwt}$. 

To quantify the convergence degree of different objectives, we further define two centralized attentive critics associated with two objective-specific optimal policies with respect to reward $r^{cwt}$ and $r^{cp}$, denoted as $Q_{\bm{b}_{cwt}^{^*}}^{cwt^*}$ and $Q_{\bm{b}_{cp}^{^*}}^{cp^*}$. 
The corresponding optimal policies of each $c^i$ are denoted as $\bm{b}_{cwt}^{^*i}$ and $\bm{b}_{cp}^{^*i}$, respectively.
Above objective-specific optimal policies and critics can be obtained by pre-training \stddpg on a single reward. 
Then, we quantify the gap ratio between multi-objective policy and objective-specific optimal policy by 
\begin{equation}\label{equ:gap_ratio}
	g_t^{cwt} = \frac{Q_{\bm{b}_{cwt}^{^*}}^{cwt^*}(x_t)|_{a_{t}^i=\bm{b}_{cwt}^{^*i}(o_{t}^i)} - Q_{\bm{b}}^{cwt}(x_t)|_{a_{t}^i=\bm{b}^{i}(o_{t}^i)}} {Q_{\bm{b}_{cwt}^{^*}}^{cwt^*}(x_t)|_{a_{t}^i=\bm{b}_{cwt}^{^*i}(o_{t}^i)}}.
\end{equation}
The gap ratio $g_t^{cp}$ can be derived in the same way.
Intuitively, a larger gap ratio indicates a poorly-optimized objective and should be reinforced with a larger update weight, while a small gap ratio indicates a well-optimized objective can be fine-tuned with a smaller step size. 
Thus, we derive dynamic update weights to adaptively adjust the step size of the two objectives, which is learned by the Boltzmann softmax function,
\begin{equation}\label{equ:update_weight}
    \beta_t = \frac{exp(g_t^{cwt}/\sigma)}{exp(g_t^{cwt}/\sigma) + exp(g_t^{cp}/\sigma)},
\end{equation}
where $\sigma$ is the temperature controls the adjustment sensitivity.

With the above two critics and adaptive update weights, the goal of each agent $c^i \in C_t^a$ is to learn an actor policy to maximize the following return,
\begin{equation}
J(\bm{b}^i)=\mathbb{E}_{s^a_t,p_t \sim D}\left[ \left(\beta_t Q_{\bm{b}}^{cwt}(x_t) + (1-\beta_t) Q_{\bm{b}}^{cp}(x_t)\right)|_{a_t^i=\bm{b}^i(o_t^i)}\right].
\end{equation}
\eat{where $\beta_t$ is the dynamic weight to adjust the optimized strength for objectives.}

\eat{Intuitively, the objective with less expected return~(\ie $Q_{\bm{b}}$) means it's
in poor-optimized and should be reinforced by a larger weight, while the well-optimized objective should be fine-turned carefully with a smaller weight. 
Thus, we leverage the boltzmann machines to quantify the weights by regarding the expected return as energy
\begin{equation}
\beta_t = \frac{exp(-Q_{\bm{b}}^{cwt}(x_t)/\sigma)}{exp(-Q_{\bm{b}}^{cwt}(x_t)/\sigma) + exp(-Q_{\bm{b}}^{cp}(x_t)/\sigma)}
\end{equation}
where $\sigma$ denotes temperature hyper-parameter to control the degree of reinforcement.}

The complete learning procedure of \stddpg is detailed in Algorithm~\ref{alg:master}. 
Note that for the consideration of scalability, we share the parameters of actor and critic networks by all agents.

\section{Experiments}\label{sec:exp}
\subsection{Experimental setup}
\subsubsection{Data description.} 
\begin{figure}[tb]
	\centering
	\includegraphics[width=0.85\columnwidth]{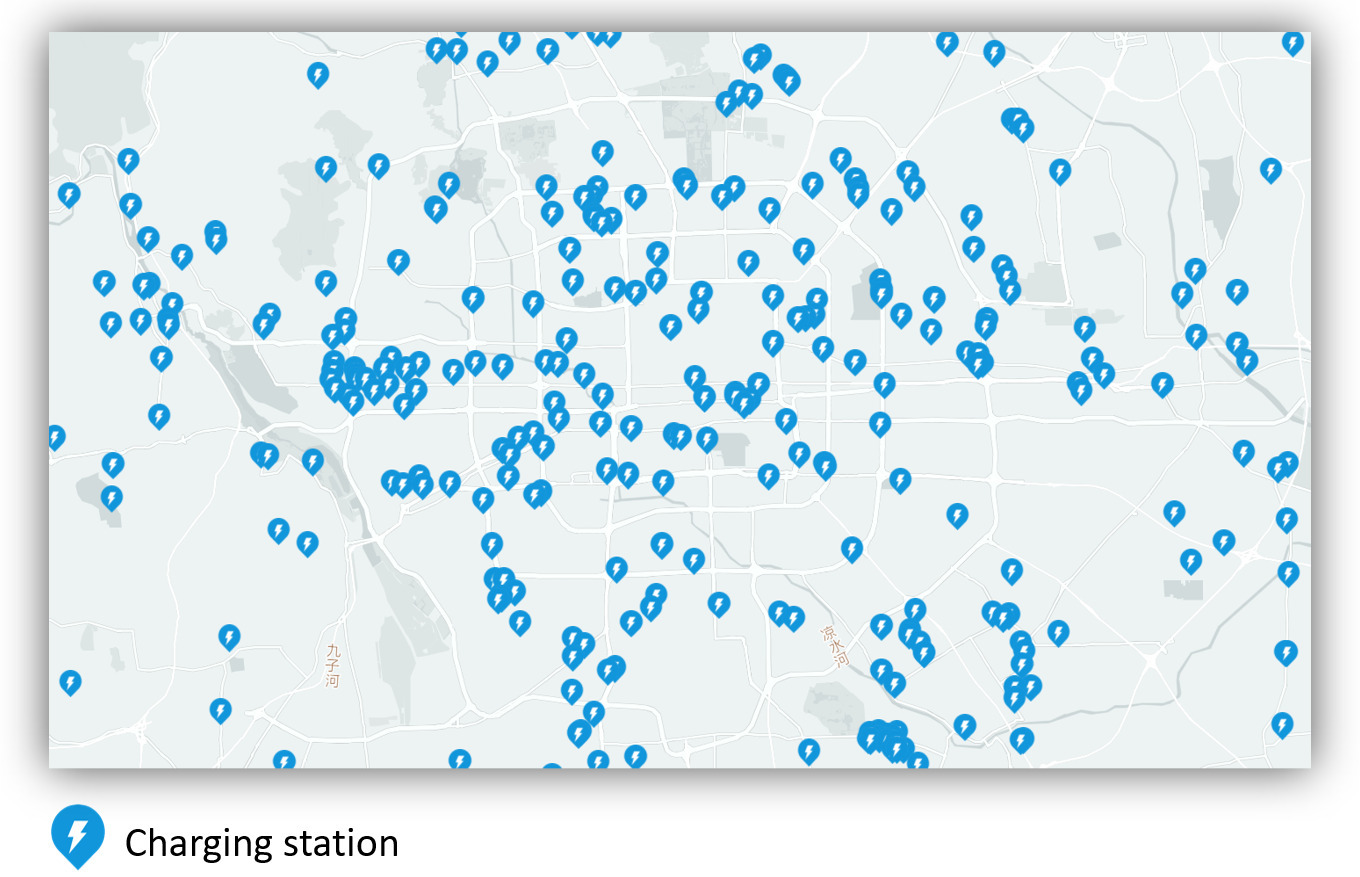}
	\caption{Spatial distribution of charging stations on \beijing.}
	\vspace{-1mm}
	\label{fig:beijing_cs}
\end{figure}

\begin{table}[htb]
	\small
	\centering
	\caption{Statistics of datasets.}
	\vspace{-2mm}
	\setlength{\tabcolsep}{3mm}{
	\begin{tabular}{lrr}
		\toprule[0.8pt]
		\textbf{Description} & \beijing & \shanghai\\
        \midrule[0.5pt]
		\# of charging stations & 596 & 367\\
		\# of supplies records & 38,620,800 & 23,781,600\\
		\# of charging requests & 152,889 & 87,142 \\
        \bottomrule[0.5pt]
	\end{tabular}}
	\label{table:dataset}
\end{table}
We evaluate \stddpg on two real-world datasets, \beijing and \shanghai, which represent two metropolises in China. Both datasets are ranged from May 18, 2019, to July 01, 2019. All real-time availability~(supplies) records, charging prices and charging powers of charging stations are crawled from a publicly accessible app~\liu{\cite{zhang2020semi_aaai}}, in which all charging spot occupancy information is collected by real-time sensors. The charging request data is collected through Baidu Maps API~\liu{\cite{hao2019trans2vec,yuan2020spatio}}. 
We split each city as $1 \times 1\text{km}^2$ grids, and aggregate the number of future $15$ minutes charging requests in the surrounding area~(\ie the grid the station locates in and eight neighboring grids) as the future demands of the corresponding charging stations. We take the first $28$ consecutive days data as the training set, the following three days data as validation set, and the rest $14$ days data for testing. 
All real-world data are loaded into an EV Charging Recommendation simulator~(refer to \appref{app:simulator} and source code for details) for simulation.
The spatial distribution of charging stations on \beijing is shown in \figref{fig:beijing_cs}.
The statistics of two datasets are summarized in \tabref{table:dataset}.
More statistical analysis results of the dataset can be found in \appref{app:data_analysis}.


\subsubsection{Implementation details.}
All experiments are performed on a Linux server with 26-core Intel(R) Xeon(R) Gold 5117 CPU @ 2:00 GHz and
NVIDIA Tesla P40 GPU. We set $d=30$ minutes for charging competition modeling, set temperature $\sigma=0.2$ for updated weights adjustment, and select discount factor $\gamma=0.99$ for learning all RL algorithms.
The actor networks and critic networks both consist of three linear network layers with dimension $64$ and ReLU activation function for hidden layers. The replay buffer size is $1000$, batch size is $32$, and we set $\tau=0.001$ for target networks soft update.
We employ Adam optimizer for all learnable algorithms and set learning rate as $5\mathrm{e}{-4}$ to train our model. We carefully tuned major hyper-parameters of each baseline via a grid search strategy. All RL algorithms are trained to recommend the top-$50$ nearest charging stations for $60$ iterations and chosen the best iteration by validation set for testing. Detailed settings of the simulator can be found in \appref{app:simulator}. 

\subsubsection{Evaluation metrics.}
We define four metrics to evaluate the performance of our approach and baseline recommendation algorithms.
Define $Q^a$ as the set of charging requests who accept our recommendations. We further define $Q^s \subseteq Q^a$ as the set of charging requests who accept our recommendations and finally succeed in charging. $|Q^a|$ and $|Q^s|$ are the cardinalities of $Q^a$ and $Q^s$, respectively.

To evaluate the overall charging wait time of our recommendations, we define Mean Charging Wait Time~(MCWT) over all charging requests $q_t\in Q^a$:
$$
\text{MCWT} = \frac{\sum_{q_t \in Q^a}CWT(q_t)}{|Q^a|},
$$
where $CWT(q_t)$ is the charging wait time~(in minute) of charging request $q_t$. 

To evaluate the average charging price, we define Mean Charging Price~(MCP) over all charging requests $q_t\in Q^s$:
$$
\text{MCP} = \frac{\sum_{q_t \in Q^s}CP(q_t)}{|Q^s|},
$$
where $CP(q_t)$ is the charging price~(in CNY) of $q_t$. 

We further define the Total Saving Fee~(TSF) to evaluate the average total saving fees per day by comparing our recommendation algorithm with the ground truth charging actions:
$$
\text{TSF} = \frac{\sum_{q_t \in Q^s}\left(RCP(q_t)-CP(q_t)\right) \times CQ(q_t)}{N_d},
$$
where $RCP(q_t)$ is the charging price of ground truth charging action and $CQ(q_t)$ is the electric charging quantity of $q_t$, and $N_d$ is the number of evaluation days. Note the TSF can be a negative number which indicates how many fees overspend comparing with the ground truth charging actions.

We finally define the Charging Failure Rate~(CFR) to evaluate the ratio of charging failures in our recommendations:
$$
\text{CFR} = \frac{|Q^a|-|Q^s|}{|Q^a|}.
$$

\subsubsection{Baselines.} 
We compare our approach with the following nine
baselines and one basic variant of \stddpg:
\begin{itemize}
	\item \textbf{Real} is the ground truth charging actions of charging requests.
	\item \textbf{Random} randomly recommends charging stations for charging.
	\item \textbf{Greedy-N} recommends the nearest charging station.
	\item \textbf{Greedy-P-5} recommends the least expensive charging station among the top-$5$ nearest stations.
	\item \textbf{Greedy-P-10} recommends the least expensive charging station among the top-$10$ nearest stations.
	\item \textbf{CDQN}~\cite{mnih2015human} is a centralized deep q-network approach, all charging stations are controlled by a centralized agent. CDQN makes recommendation based on the state of all charging stations. The action-value function is a $3$ layers MLP with dimension $256$ and ReLU activation for hidden layers. The replay buffer size is $2000$ and batch size is $64$. The learning rate is set to $1\mathrm{e}{-3}$ and we use the decayed $\epsilon$-greedy strategy for exploration.
	\item \textbf{CPPO}~\cite{schulman2017proximal} is a centralized policy gradient approach, the recommendation mode is the same as CDQN. The policy network and value function network both consist of $3$ layers MLP with dimension $256$ and ReLU activation for hidden layers. The learning rate for policy and value networks are both set to $5\mathrm{e}{-4}$, the $\epsilon$ for clipping probability ratio is $0.2$.
	\item \textbf{IDDPG}~\cite{lillicrap2015continuous} is a straight-forward approach to achieve MARL using DDPG, where all agents are completely independent. The critic network
	approximates the expected return only bases on agent-specific observation and action. Other hyper-parameter settings are the same as \stddpg.
	\item \textbf{MADDPG}~\cite{lowe2017multi} is a state-of-the-art algorithm for cooperative MARL. The actor takes action based on agent-specific observation, but the critic can access full state and joint action in training. The critic consists of $3$ layers MLP with dimension $256$ and ReLU activation for hidden layers. Other hyper-parameter settings are the same as \stddpg.
	For scaling MADDPG to the large-scale agent system, we share actor and critic networks among all agents. 
	\item \textbf{MASTER-ATT} is a basic variant of \stddpg without charging competition information and multi-critic architecture. 
\end{itemize}

\begin{table}[htb]
	\centering
	\caption{Overall performance evaluated by MCWT, MCP, TSF and CFR on \beijing.}
	\vspace{-2mm}
	\begin{tabular}{lcccc}
		\toprule[0.8pt]
		\multirow{2}{*}{\textbf{Algorithm}} & \multicolumn{4}{c}{\beijing} \\ \cline{2-5}
		& MCWT & MCP & TSF & CFR \\ 
        \midrule[0.5pt]
		Real &  21.51 & 1.749 & - & 25.9\%  \\
		Random &  38.77 & 1.756 & -447 & 52.9\% \\
		Greedy-N &  20.27 & 1.791 & -2527 & 31.3\% \\
		Greedy-P-5 &  23.40 & 1.541 & 9701 & 35.4\% \\
		Greedy-P-10 &  26.03 & \textbf{1.424} & 14059 & 39.9\% \\
		CDQN &  19.24 & 1.598 & 9683 & 7.4\% \\
		CPPO &  17.67 & 1.639 & 6707 & 5.3\% \\
		IDDPG &  13.26 & 1.583 & 11349 & 2.2\%\\
		MADDPG &  14.01 & 1.570 & 12033 & 4.7\% \\
        \midrule[0.5pt]
		\stddpg-ATT & 11.30 & 1.562 & 12661 & 1.8\% \\
		\stddpg & \textbf{10.46} & 1.512 & \textbf{16219} & \textbf{0.9\%} \\
        \bottomrule[0.5pt]
	\end{tabular}
	\label{table:overall_beijing}
\end{table}

\begin{table}[htb]
	\centering
	\caption{Overall performance evaluated by MCWT, MCP, TSF and CFR on \shanghai.}
	\vspace{-2mm}
	\begin{tabular}{lcccc}
		\toprule[0.8pt]
		\multirow{2}{*}{\textbf{Algorithm}} & \multicolumn{4}{c}{\shanghai} \\ \cline{2-5}
		& MCWT & MCP & TSF & CFR \\ 
		\midrule[0.5pt]
		Real &   19.31 & 1.787 & - & 16.7\% \\
		Random &   38.62 & 1.826 & -698 & 46.1\% \\
		Greedy-N &  14.44 & 1.838 & -1252 & 16.9\%  \\
		Greedy-P-5 &  16.65 & 1.502 & 10842 & 13.8\% \\
		Greedy-P-10 & 19.60 & \textbf{1.357} & \textbf{15649} & 16.5\% \\
		CDQN & 20.74 & 1.686 & 4650 & 6.5\% \\
		CPPO & 18.84 & 1.750 & 1918 & 4.6\%\\
		IDDPG & 14.02 & 1.562 & 9720 & 2.6\%\\
		MADDPG & 13.63 & 1.553 & 10209 & 2.8\%\\
        \midrule[0.5pt]
		\stddpg-ATT & 12.24 & 1.548 & 10630 & 2.2\%\\
		\stddpg & \textbf{11.80} & 1.497 & 12497 & \textbf{1.5\%}\\
		\bottomrule[0.5pt]
	\end{tabular}
	\label{table:overall_shanghai}
\end{table}


\subsection{Overall Performance}
\tabref{table:overall_beijing} and \tabref{table:overall_shanghai} report the overall results of our methods and all the compared baselines on two datasets with respect to our four metrics.
As can be seen, overall, \stddpg achieves the most well-rounded performance among all the baselines.
Specifically, \stddpg reduces $(51.4\%, 13.6\%, 96.5\%)$ and $(38.9\%, 16.2\%, 91.0\%)$ for (MCWT, MCP, CFR) compared with the ground truth charging actions on \beijing and \shanghai, respectively. And through our recommendation algorithm, we totally help users saving $16,219$ and $12,497$ charging fees per day on \beijing and \shanghai.
Besides, \stddpg achieves $(21.1\%, 4.5\%, 42.9\%, 59.1\%)$ and $(25.3\%,\\ 3.7\%, 34.8\%, 80.9\%)$ improvements for (MCWT, MCP, TSF, CFR) compared with two multi-agent baselines IDDPG and MADDPG on \beijing , and the improvements are $(15.8\%, 4.2\%, 28.6\%, 42.3\%)$ and $(13.4\%, 3.6\%, 22.4\%, 46.4\%)$ on \shanghai. 
All the above results demonstrate the effectiveness of \stddpg.
Look into \stddpg-ATT and MADDPG, two MARL algorithms with centralized training decentralized execution. We can observe the \stddpg-ATT with centralized attentive critic have all-sided improvements comparing with MADDPG, especially in the MCWT and CFR. This is because the critic in MADDPG integrates full state and joint action suffering from the large state and action space problem, while our centralized attentive critic can effectively avoid this problem in our task.

Looking further into the results, we observe centralized RL algorithms~(\ie CDQN and CPPO) consistently perform worse than multi-agent approaches~(\ie IDDPG, MADDPG, \stddpg-ATT and \stddpg), which validates our intuition that the centralized algorithms suffer from the high dimensional state and action space problem and hard to learn satisfying policies, while multi-agent approaches perform better because of their quantity-independent state action space in large-scale agents environment. 
We also observe that Greedy-P-10 has the best performance in MCP among all compared algorithms. This is no surprise, because it always recommends the least expensive charging stations without considering other goals, which causes that the Greedy-P-10 performs badly in MCWT and CFR. It is worth mentioning that \stddpg unexpectedly exceeds Greedy-P-10 in TSF on \beijing for the reason that \stddpg has a much lower CFR comparing with Greedy-P-10, more users are benefited from our recommendation.

\subsection{Ablation Study}
\begin{figure}[tb]
\centering
\subfigure[{\small MCWT}]{
	\includegraphics[width=0.48\columnwidth]{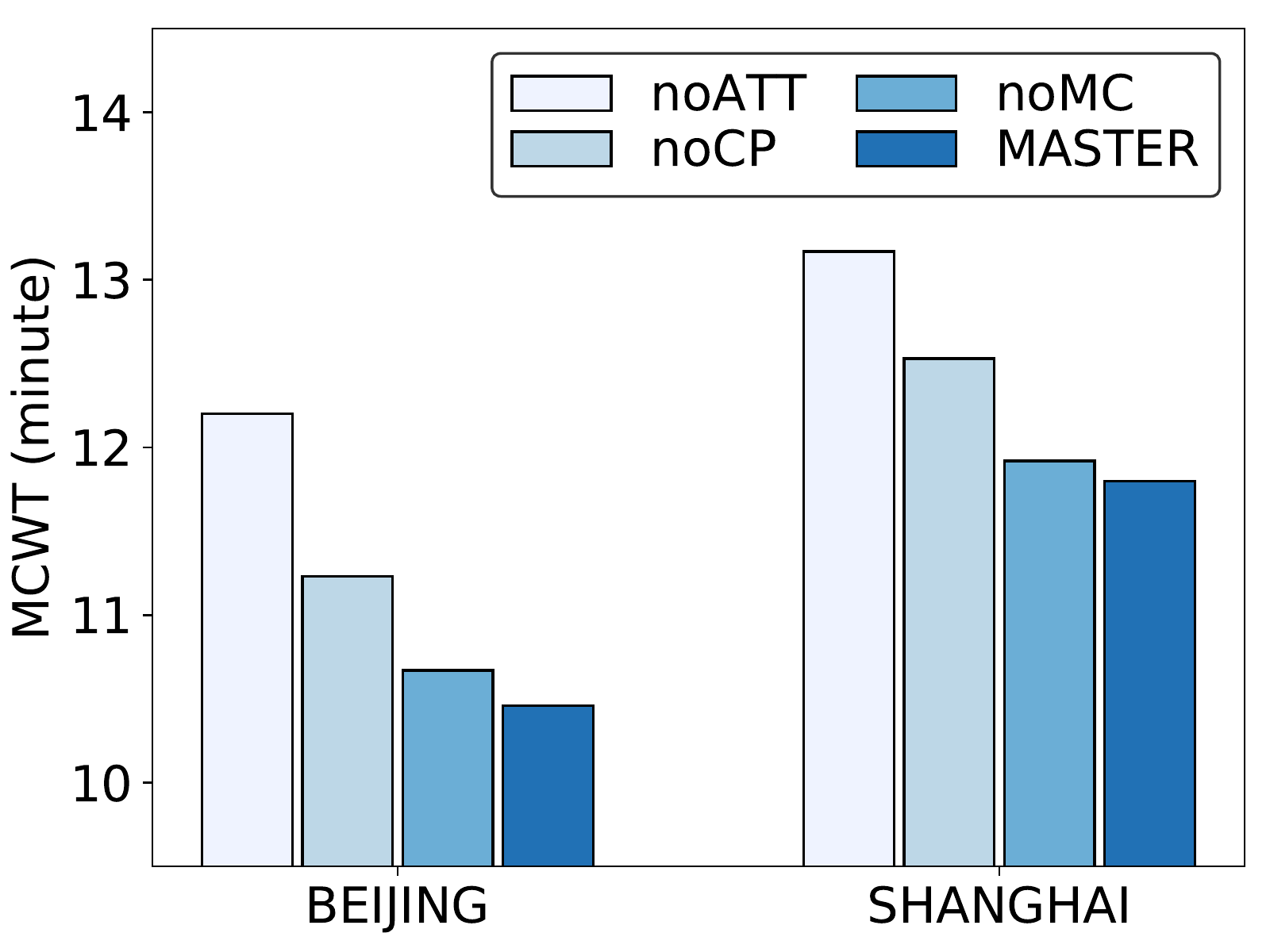}}
\subfigure[{\small MCP}]{
	\includegraphics[width=0.48\columnwidth]{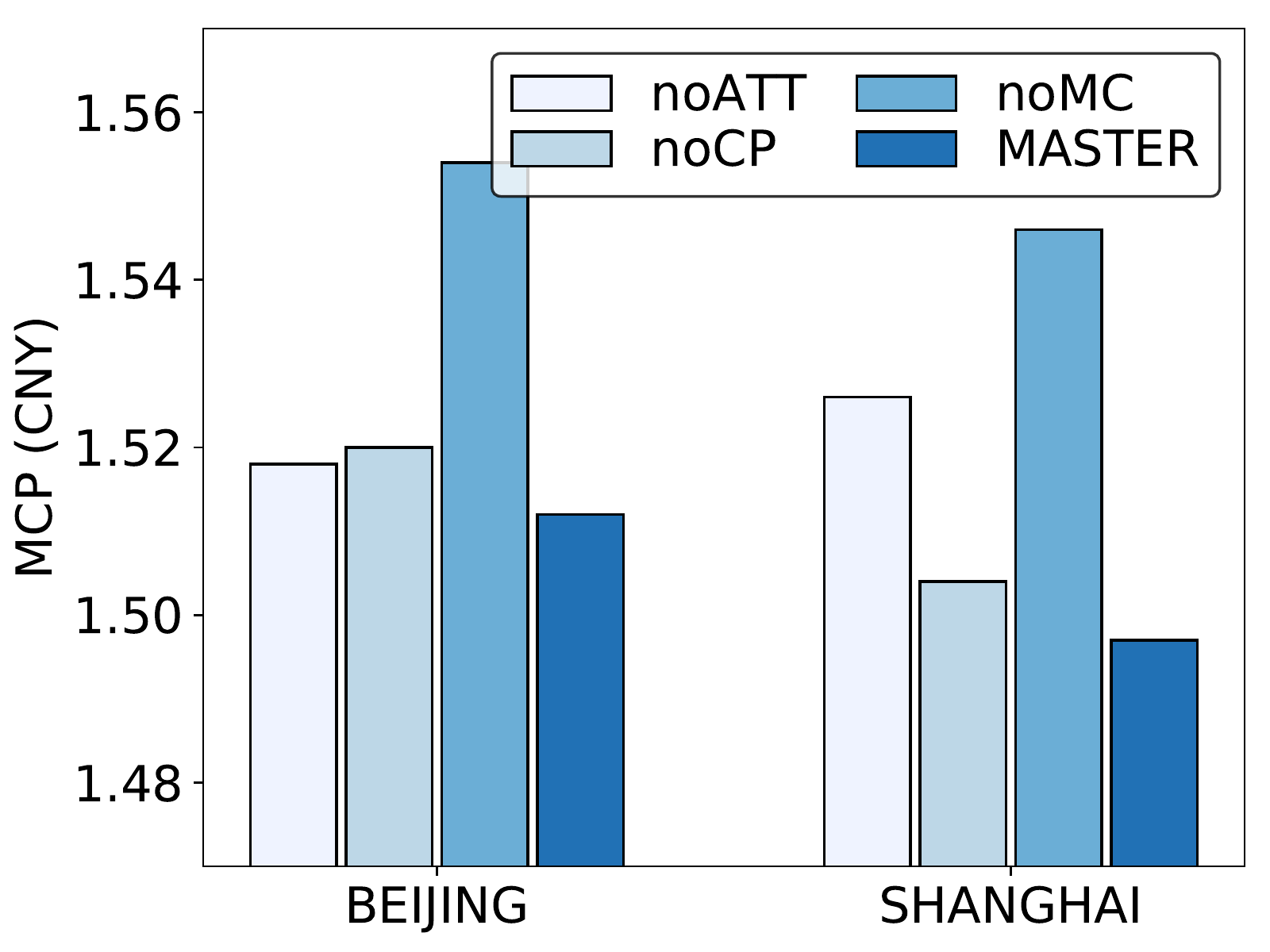}}
\subfigure[{\small TSF}]{
    \includegraphics[width=0.485\columnwidth]{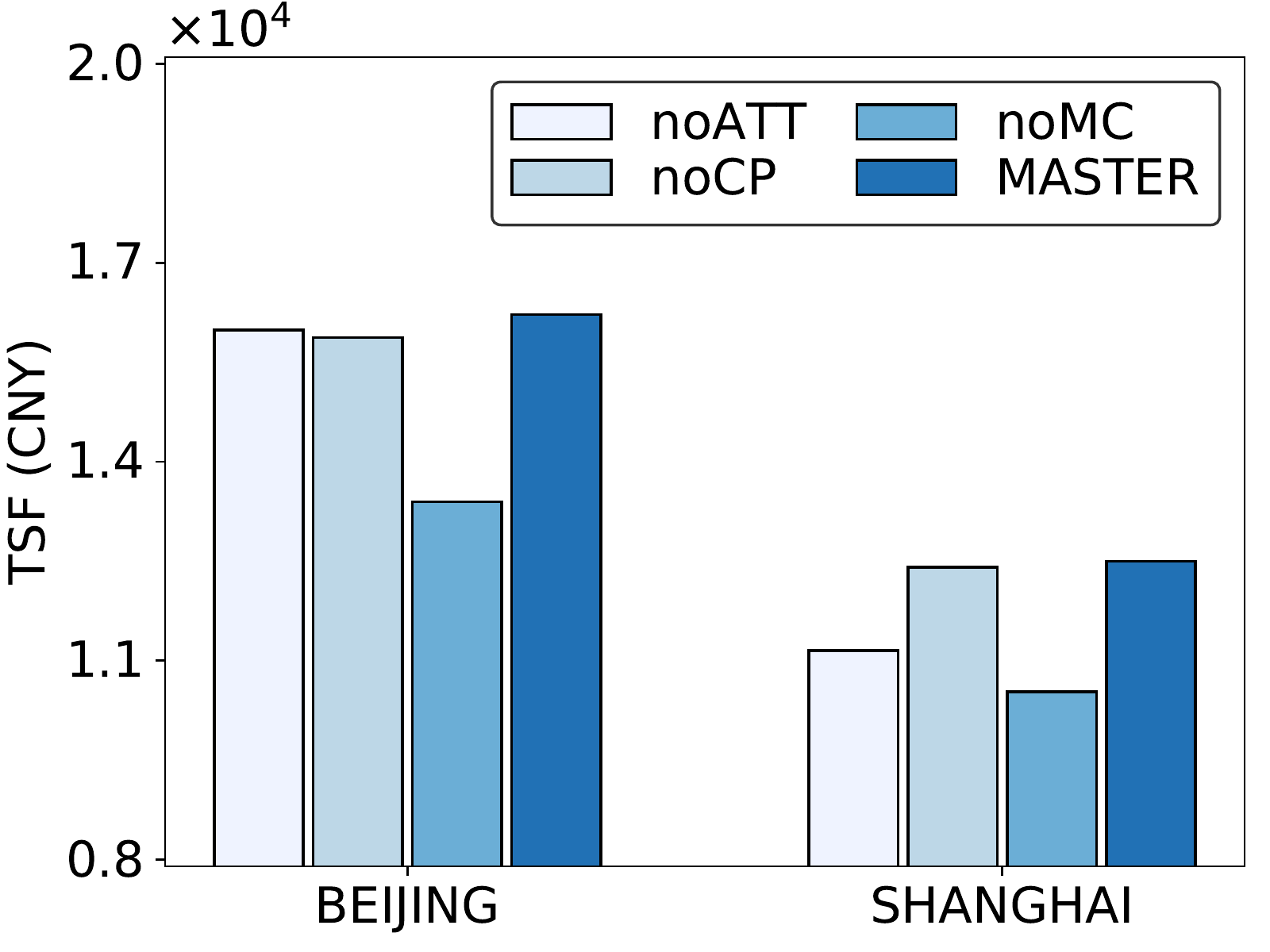}}
\subfigure[{\small CFR}]{
	\includegraphics[width=0.48\columnwidth]{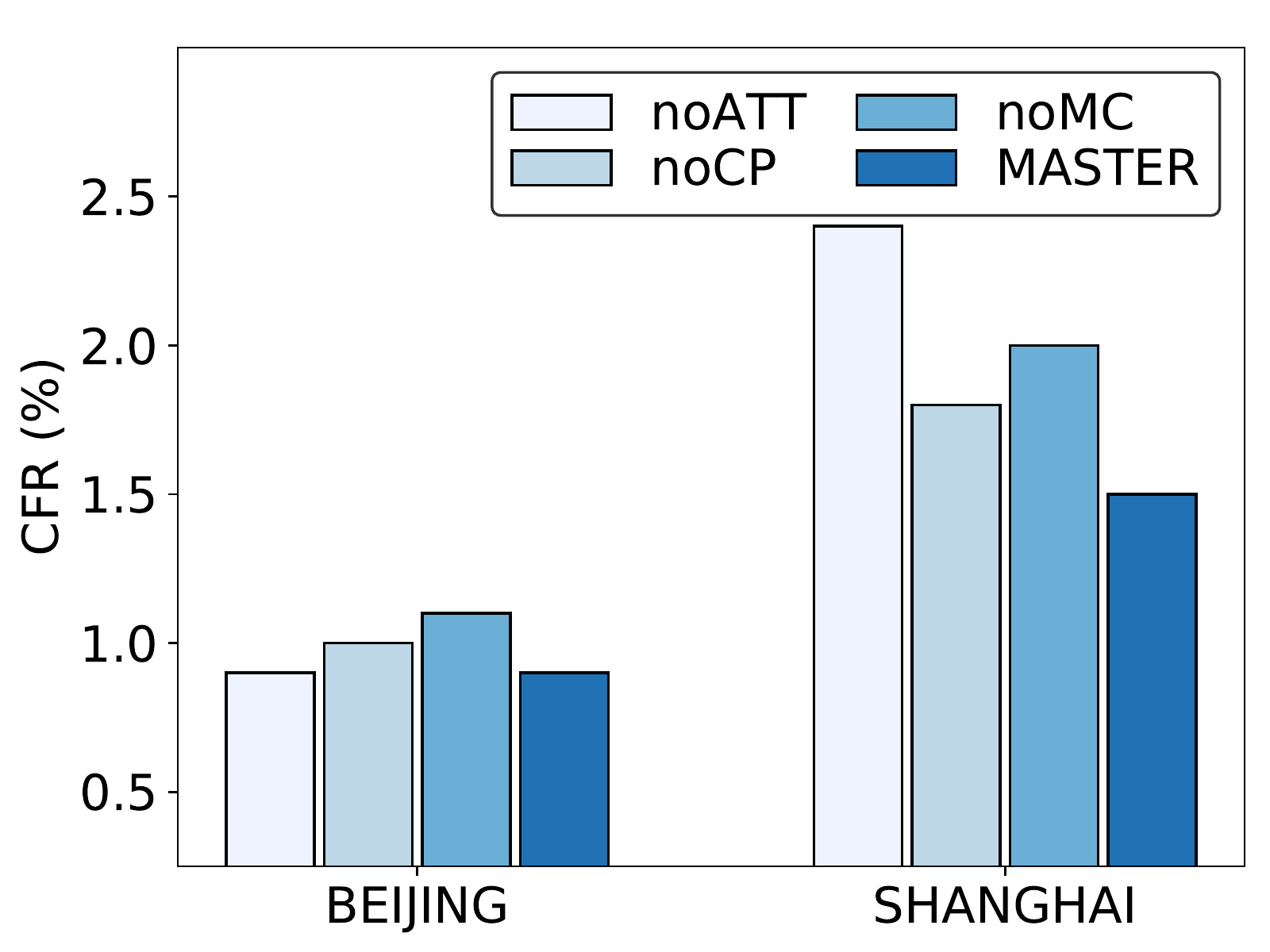}}
\caption{Ablation studies for MCWT, MCP, TSF and CFR on \beijing and \shanghai}
\label{exp:ablation_study}
\end{figure} 
In this section, we conduct ablation studies on \stddpg to further verify the effectiveness of each component. We evaluate the performance of \stddpg and it's three variants for the four metrics on \beijing and \shanghai. Specifically, the three variants are (1)~\emph{noATT} removes centralized attentive critic, learning agents independently;
(2)~\emph{noCP} removes the future potential charging competition information from centralized attentive critic;
(3)~\emph{noMC} removes multi-critic architecture, learning policies by a single centralized attentive critic with the average combined reward. The ablation results are reported in \figref{exp:ablation_study}. As can be seen, removing centralized attentive critic~(\emph{noATT}) or charging competition information~(\emph{noCP}) have a greater impact on MCWT. This is because MCWT is highly related to the cooperation of agents and charging competition of EVs, whereas the removed modules exactly work on these two aspects. Removing multi-critic architecture~(\emph{noMC}) has a large performance degradation on MCP and TSF. This is because \emph{noMC} suffers from the convergence problem and finally leads to performance variance on multiple objectives.
All the above experimental results demonstrate the effectiveness of each component in \stddpg.

\subsection{Effect of Recommendation Candidates}
\begin{figure}[tb]
\centering
\vspace{-1.25mm}
\subfigure[{\small MCWT and MCP}]{\label{exp:mcwt_mcp}
	\includegraphics[width=0.48\columnwidth]{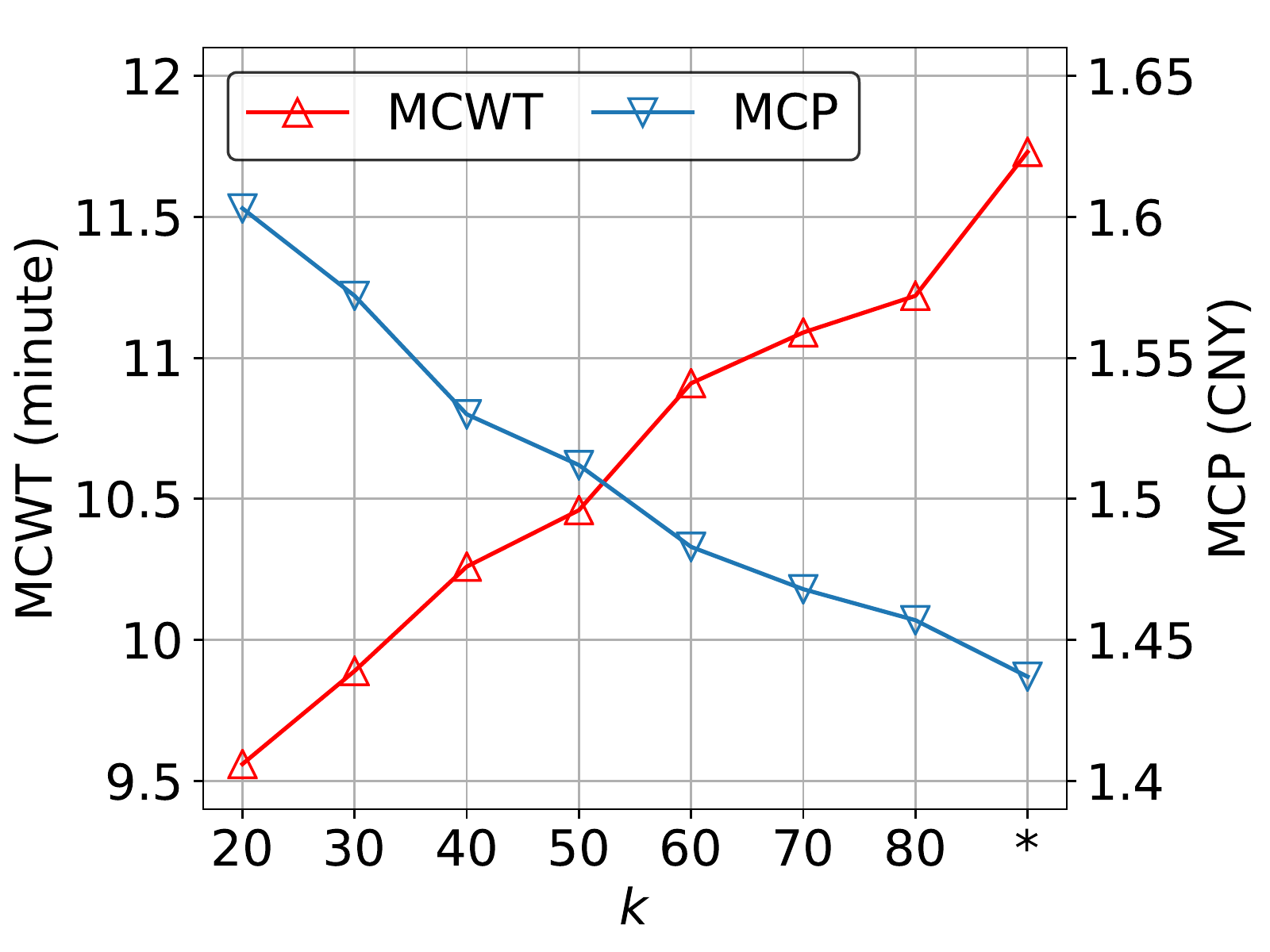}}
\subfigure[{\small TSF and CFR}]{\label{exp:tsf_CFR}
	\includegraphics[width=0.485\columnwidth]{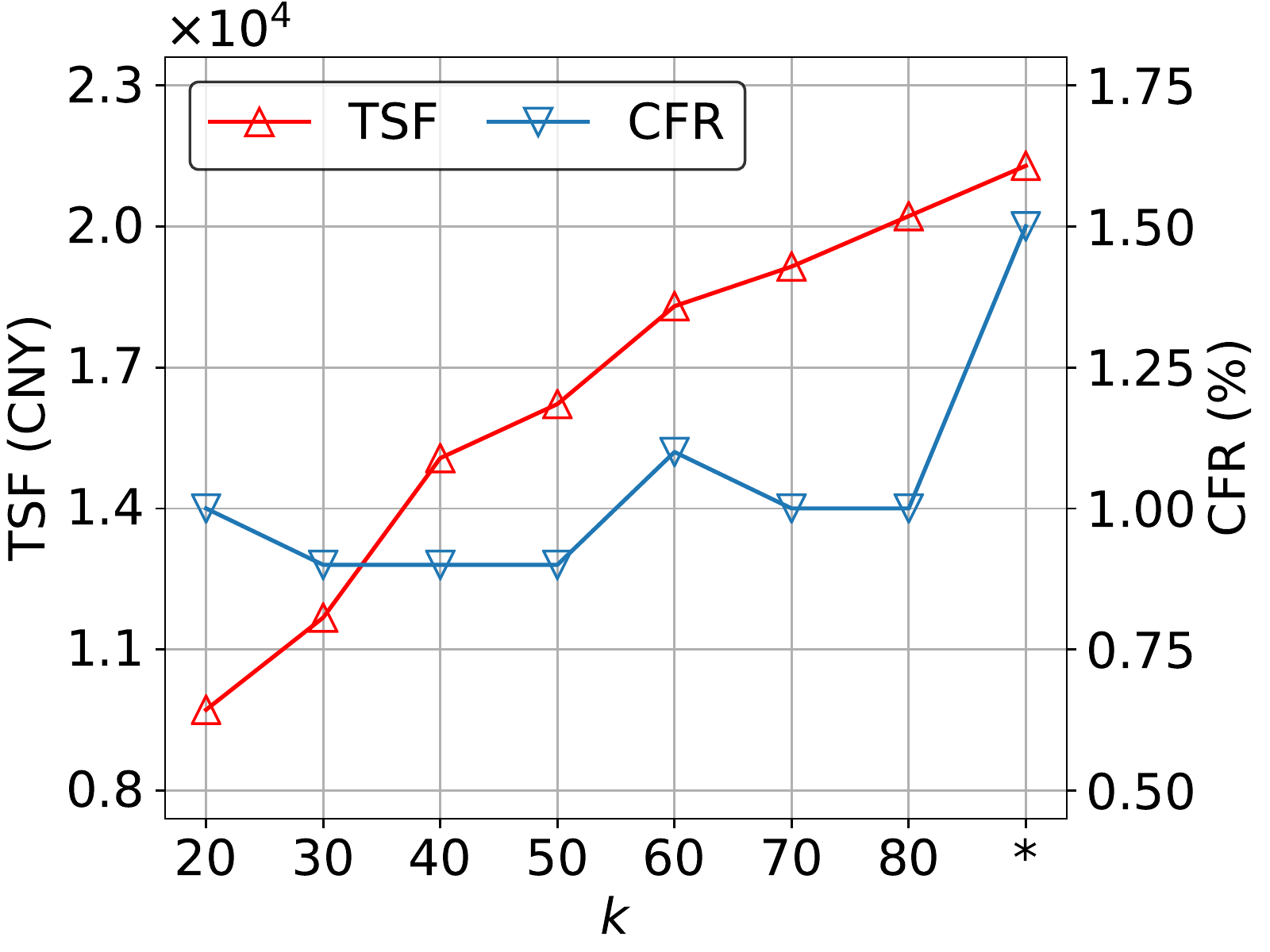}}
\caption{Effect of top-$k$ nearest charging stations for recommendation performance.}
\label{exp:topk}
\end{figure} 

In this section, we study the effect of the number of top-$k$ active agents for recommendations on \beijing. We vary $k$ from $20$ to the number of all the agents. The results are reported in \figref{exp:topk}, where "$*$" denotes all the agents are activated for recommendation decision. As can be seen, with relaxing the recommendation constraint $k$, the MCWT increases while the MCP and TSF decrease, further validating the divergence of the optimal solutions for different objectives.
This is possible because a more relaxed candidate number will expose more alternative economic but distant charging stations for recommendations. 
Even so, we should notice that the performance under the most strict constraint~(\ie $k=20$) or without constraint~(\ie $k=*$) are varying in a small range, \ie the (MCWT, MCP) are ($9.56, 1.603$) and ($11.73, 1.437$), respectively, are not extreme and still acceptable for the online recommendation. The above results demonstrate that our model are well-rounded with different candidate numbers.
It also inspires us that we can explore to make diversified recommendations that are biased to different objectives to satisfy personalized preferences in the future.

\subsection{Convergence Analysis}
\begin{figure}[t]
\centering
\subfigure[{\small Static updated weight}]{\label{exp:static}
	\includegraphics[width=0.48\columnwidth]{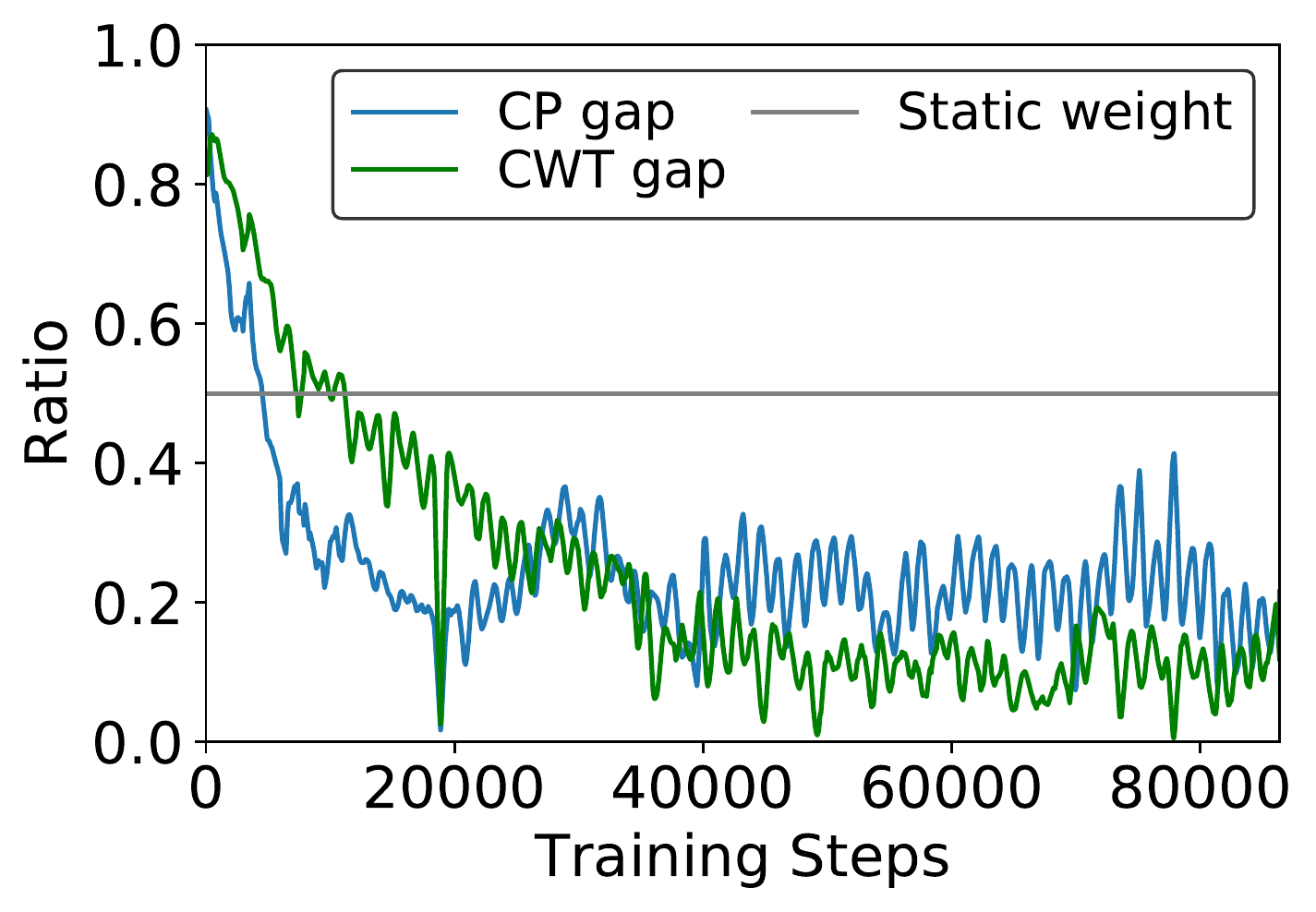}}
\subfigure[{\small Adaptive updated weight}]{\label{exp:dynamic}
	\includegraphics[width=0.48\columnwidth]{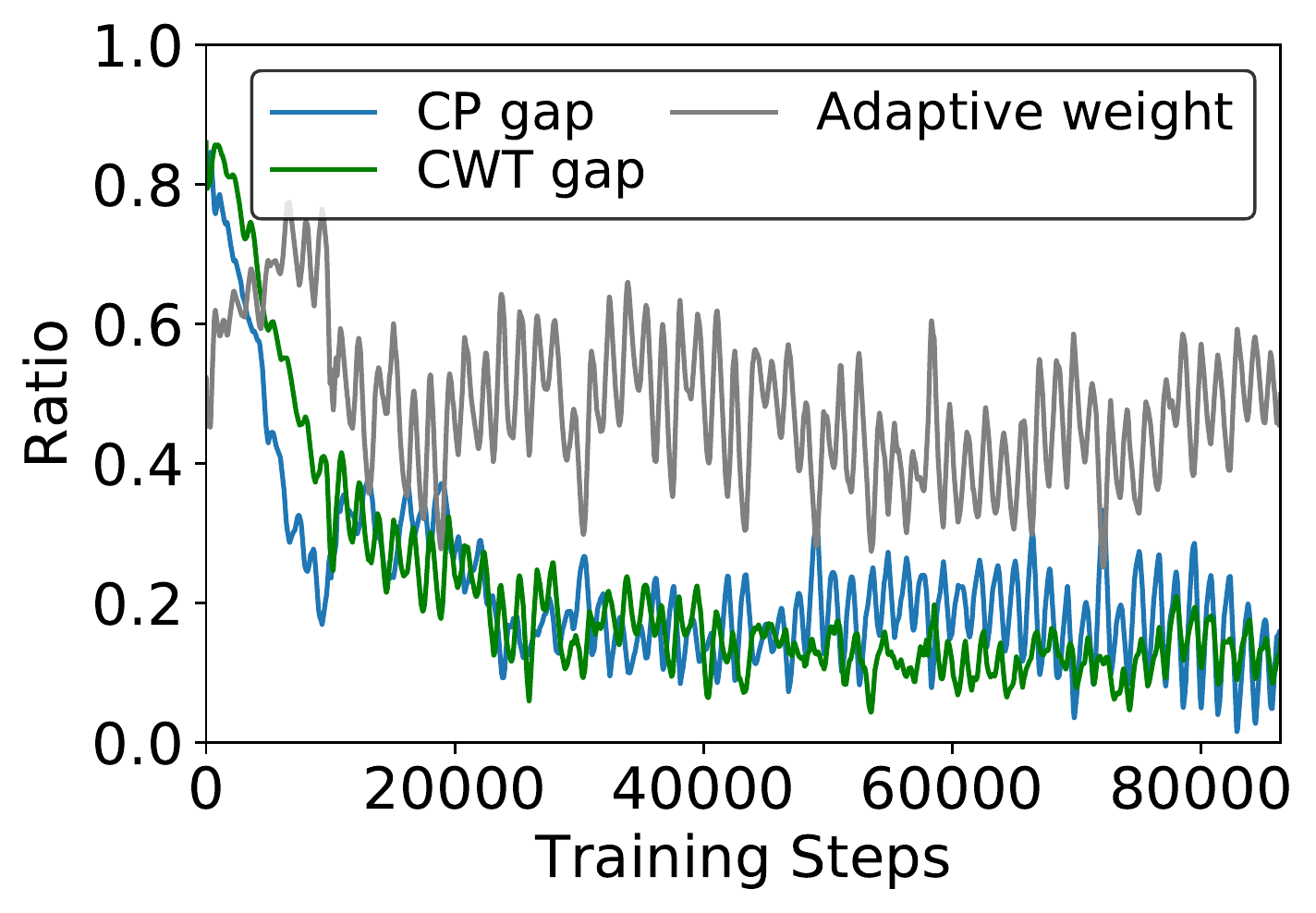}}
\caption{Convergence curves by moving average for every 500 successive training steps.}
 \vspace{-3mm}
\label{exp:convergence}
\end{figure} 
In this section, we study the effect of our multi-objective optimization method by comparing the convergence of \stddpg between the static and our adaptive strategy. Note the \stddpg with static average updated weight is equivalent to \stddpg-noMC~(\ie average two rewards as a single reward for training). More details can be found in proposition \ref{pf:avg_reward} in the \appref{app:prop}. 
The gap ratios~(\equref{equ:gap_ratio}) and dynamic weights~(\equref{equ:update_weight}) are reported in \figref{exp:convergence}. As can be seen in \figref{exp:static}, at the beginning of training, the gap ratio of the CP critic is decreasing faster than the CWT one. This is because the distribution of CP reward is more concentrated ~(as illustrated in \figref{fig:reward_dist}), it's easier to accumulate rewards comparing with CWT when agents with poor policies.
In such a case, the static method keeps invariant update weights and results in slow convergence for CWT. 
In contrast, as shown in \figref{exp:dynamic}, the dynamic re-weighting method in \stddpg assigns a higher weight to the CWT objective when it falls behind the optimization of CP, and adjust them to a synchronous convergence pace.
As the training step increase, the convergence of CWT overtakes the CP. This is because the distribution of CWT reward is more even, having
larger room for improvement.
And the optimization of CWT drags the CP objective to a sub-optimal point since the divergences between two objectives.
However, the static method is helpless for
such asynchronous convergence, but \stddpg with the dynamic gradient re-weighting strategy weakens such asynchronism, and finally adjust them to be synchronous~(\ie~weight approaches 0.5) to learn well-rounded policies.


\subsection{Case Study}
\begin{figure}[t]
\centering
\subfigure[{\small Case 1}]{
	\includegraphics[width=0.56\columnwidth]{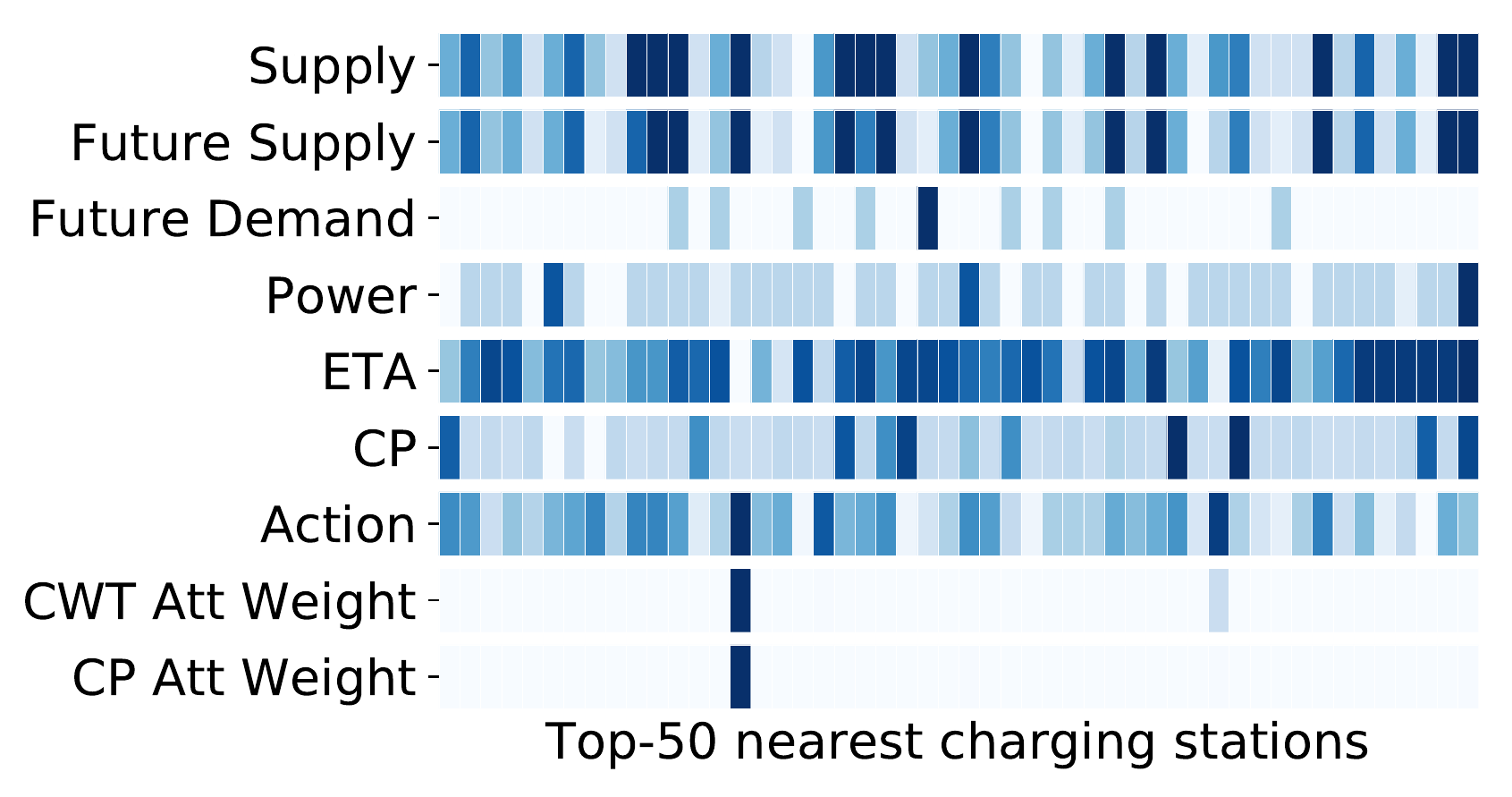}}
\subfigure[{\small Case 2}]{
	\includegraphics[width=0.41\columnwidth]{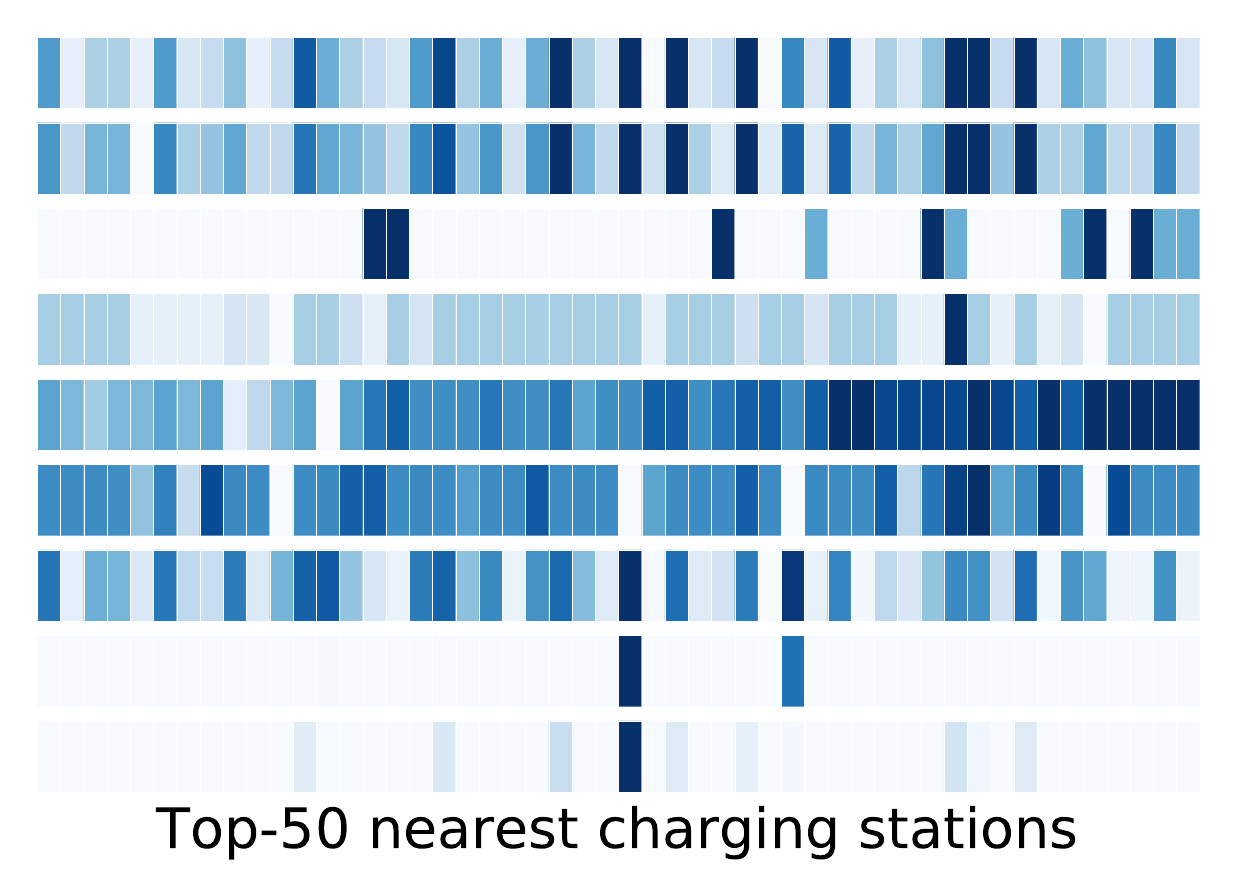}}
\caption{Cases of the centralized attentive critics. For each factor~(\ie each row), the darker color means a greater value. And each col denotes a top-50 nearest charging station for the charging request.}
\label{fig:case_attention}
\end{figure}

In this section, we visualize the correlation among the attention weights and some input features of two centralized attentive critics~(\ie $Q_{\bm{b}}^{cwt}$ and  $Q_{\bm{b}}^{cp}$), to qualitatively analysis the effectiveness of \stddpg.
Two cases are depicted in \figref{fig:case_attention}. We can observe these two critics pay more attention to the charging stations with high action values. This makes sense, since these charging stations are highly competitive bidding participants.
The charging request will be recommended to the charging station with the highest action, then environment returns rewards depending on this recommended station.
Furthermore, we observe the action is highly correlated with factors such as supply, future supply, future demand, ETA and CP. A charging station with low ETA, CP and sufficient available charging spots~(supplies) always has a high action value. Conversely, the charging stations with few available charging spots but high future demands usually derive a low action value for avoiding future charging competitions. The above observations validate the effectiveness of our model to mediate the contradiction between the space-constrained charging capacity and spatiotemporally unbalanced charging demands.

\section{Related work}\label{sec:related}
\noindent \textbf{Charging Station Recommendation.}
In recent years, EVs have become an emerging choice in the modern transportation system due to their low-carbon emission and energy efficiency. Some efforts~\cite{cao2019toward,guo2017recommendation,kong2016line,tian2016real,wang2018bcharge,wang2019tcharge,wang2020faircharge,wang2020joint,yuan2019p} have been made for charging station recommendation for EVs.
In particular, most studies~\cite{wang2020faircharge,guo2017recommendation,tian2016real,wang2019tcharge,cao2019toward} focus on recommending charging station locations for EV drivers with the goal of time.
For example, \citet{guo2017recommendation} propose to make recommendations of charging stations to minimize the time of traveling and queuing with a game-theoretical approach.
\citet{wang2020faircharge} devise a fairness-aware recommender system to reduce the idle time based on the fairness constraints.
\citet{cao2019toward} introduce the charging reservation information into the vehicle-to-vehicle system to facilitate the location recommendation. 
Different from charging location recommendation problem, another line of works~\cite{yuan2019p,kong2016line,wang2018bcharge,wang2020joint} investigate to handle more complicated scenarios, especially considering commercial benefits.
\citet{yuan2019p} propose a charging strategy that allows an electric taxi to get partially charged to meet the dynamic passenger demand.
In particular, with the help of deep reinforcement learning which has been widely adopted for solving sequential decision-making problems, \citet{wang2020joint} design a multi-agent mean field hierarchical reinforcement learning framework by regarding each electric taxi as an individual agent to provide charging and relocation recommendations for electric taxi drivers, so as to maximize the cumulative rewards of the number of served orders. However, formulating each EV driver as an agent is not suitable for our problem, since most charging requests of a day in our task are ad-hoc and from non-repetitive drivers.
Besides, the above works mainly focus on a single recommendation objective, which can not handle multiple divergent recommendation objectives simultaneously.

\noindent \textbf{Multi-Agent Reinforcement Learning.}
Multi-agent Reinforcement Learning~(MARL) is an emerging sub-field of reinforcement learning. Compared with traditional reinforcement learning, MARL expects the agents to learn to cooperate and compete with others. The simplest approach to realize a multi-agent system is learning agents independently \cite{tan1993multi,tampuu2017multiagent,gupta2017cooperative}. However, the independent agents are not able to coordinate their actions, failing to achieve complicated cooperation\cite{sukhbaatar2016learning,DBLP:journals/ker/MatignonLF12,lowe2017multi}.
A kind of natural approaches to achieve agents' cooperation is to learn communication among multiple agents \cite{foerster2016learning,sukhbaatar2016learning,peng2017multiagent,jiang2018learning}. 
However, such approaches always lead to high communication overhead because of the large amount of information transfer.
Alternatively, there are some works that employ the centralized training decentralized execution architecture to achieve agents' coordination and cooperation~\cite{lowe2017multi,DBLP:conf/aaai/FoersterFANW18,iqbal2019actor}. 
The advantage of such methods is that the agents can make decentralized execution without involving any other agents' information, which is lightweight and fault-tolerant in a large-scale agent system.


\noindent\textbf{Multi-Agent Transportation Systems.}
In the past years, a few studies have successfully applied MARL for several intelligent transportation tasks. For example, \cite{wei2019colight,wang2020stmarl} use MARL algorithm for cooperative traffic signal control. \cite{li2019efficient,lin2018efficient,jin2019coride,zhou2019multi} apply MARL into the large-scale ride-hailing system to maximize the long-term benefits. Besides, MARL also has been adopted for shared-bike repositioning \cite{li2018dynamic}, express delivery-service scheduling~\cite{liyexin2019efficient}.
However, we argue that our problem is inherently different from the above applications as a recommendation task, and the above approaches cannot be directly adopted for our problem.

\section{Conclusion}\label{sec:conclusion}
In this paper, we investigated the intelligent EV charging recommendation task with the long-term goals of simultaneously minimizing the overall CWT, average CP and the CFR. We formulated this problem as a multi-objective MARL task and proposed a spatiotemporal MARL framework, \stddpg. Specifically, by regarding each charging station as an individual agent, we developed the multi-agent actor-critic framework with centralized attentive critic to stimulate the agents to learn coordinated and cooperative policies. Besides, to enhance the recommendation effectiveness, we proposed a delayed access strategy to integrate the future charging competition information during model training. Moreover, we extend the centralized attentive critic to multi-critics with a dynamic gradient re-weighting strategy to adaptively guide the optimization direction of multiple divergent recommendation objectives. Extensive experiments on two real-world datasets demonstrated the effectiveness of \stddpg compared with nine baselines. 

\begin{acks}
This research is supported in part by grants from the National Natural Science Foundation of China (Grant No.91746301, 71531001, 62072423).
\end{acks}

\bibliographystyle{ACM-Reference-Format}
\bibliography{main}

\appendix
\section{Simulator design}
\label{app:simulator}
We develop a simulator\footnotemark[1] based on historical real-time availability~(supplies) records data, historical real-time charging prices data, and charging powers data of charging stations, along with historical electric vehicles~(EVs) charging data, and road network data~\liu{\cite{liu2020multi}}, to simulate how the system runs for a day. We take Baidu API\footnotemark[2] to compute ETA~\liu{\cite{liu2020incorporating}} between the location of charging request and charging station. 
Besides, we train a MLP model to predict the future demands based on the historical charging requests data, and leverage the learned MLP model to predict the future demands in evaluation phase.

\footnotetext[1]{https://github.com/Vvrep/MASTER-electric\_vehicle\_charging\_recommendation}
\footnotetext[2]{http://lbsyun.baidu.com/index.php?title=webapi}

\noindent \textbf{Simulation step.} Initially, our simulator loads basic supplies, charging prices and charging powers data of charging stations and charging requests by minutes of a day from the real-world dataset. For each minute $T$ in simulator, there are three events to be disposed:
\begin{itemize}
	\item \textbf{Electric vehicles' departure and arrival}: In this event, we dispose EVs' departure and arrival at $T$. For EV leaving station, the charging stations will free corresponding number of charging spots. For an arrived EV, the vehicle successfully charges if there still have available spots in the charging station, and the environment will return a negative charging wait time~(CWT) and negative charging price~(CP) as the rewards if the arrived EV accepted our recommendation.
	For each successful charging request, the vehicle will block one charging spot in a duration which obeys a Gaussian distribution associated to the charging power of each charging station.
	If the charging station is full, this EV has to queue until there have an available charging spot, or fail to charge when the CWT exceeds a predefined threshold~($45$ minutes). The penal rewards $\epsilon_{cwt}$ and $\epsilon_{cp}$ of failed request are $-60$ and $-2.8$, respectively.
	In the implementation, the number of available charging spots can be a negative number, indicating how many EVs are queuing at station for charging.
	
	\item \textbf{Charging requests recommendation}: The policy will make recommendations for each charging request at $T$. The EV of the charging request will drive to the recommended charging station based on the real recommendation acceptance probability, and otherwise will go to the ground truth charging station.
	
	\item \textbf{Transition collection}: If there have charging requests at $T$, the transitions $(o_{t'}^i,a_{t'}^i,o_{t}^i,R_{t':t})$ for charging stations will be collected and stored into replay memory \cite{mnih2015human} for learning RL algorithms.
\end{itemize}

\begin{figure}[htb]
\centering
\subfigure[{\small Charging requests}]{
	\includegraphics[width=0.49\columnwidth]{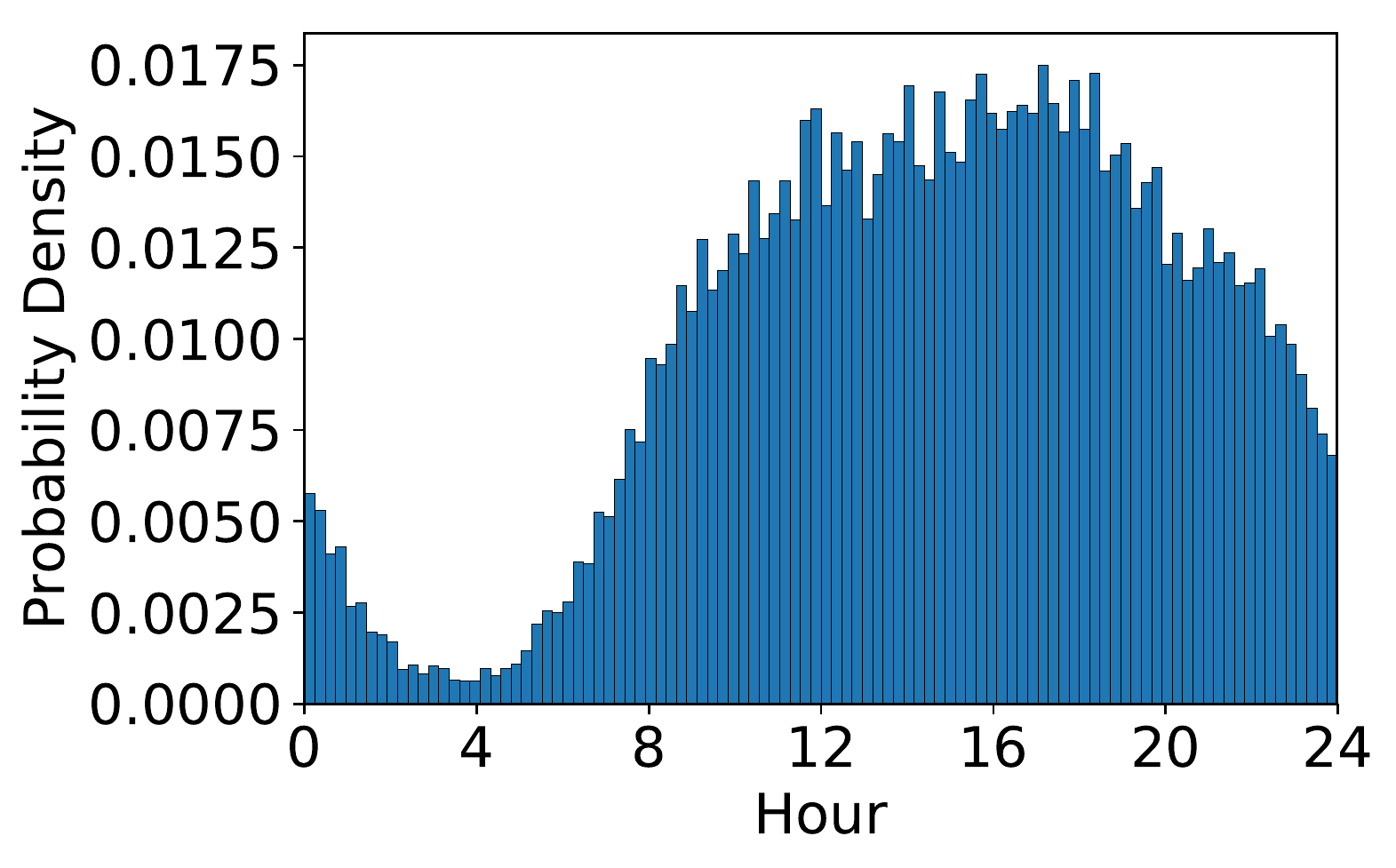}}
\subfigure[{\small ETA}]{
    \includegraphics[width=0.48\columnwidth]{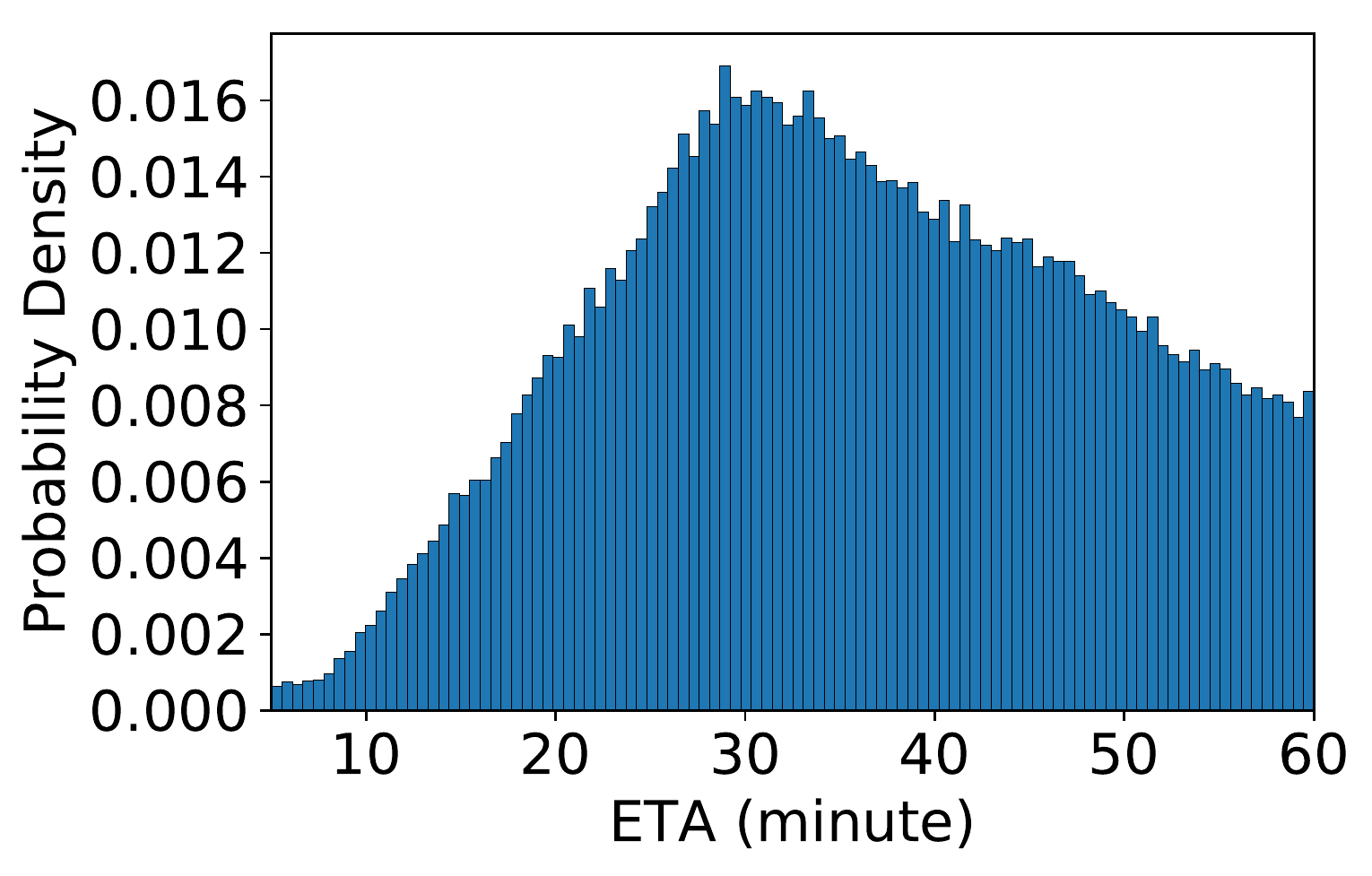}}
\subfigure[{\small Charging powers}]{
	\includegraphics[width=0.48\columnwidth]{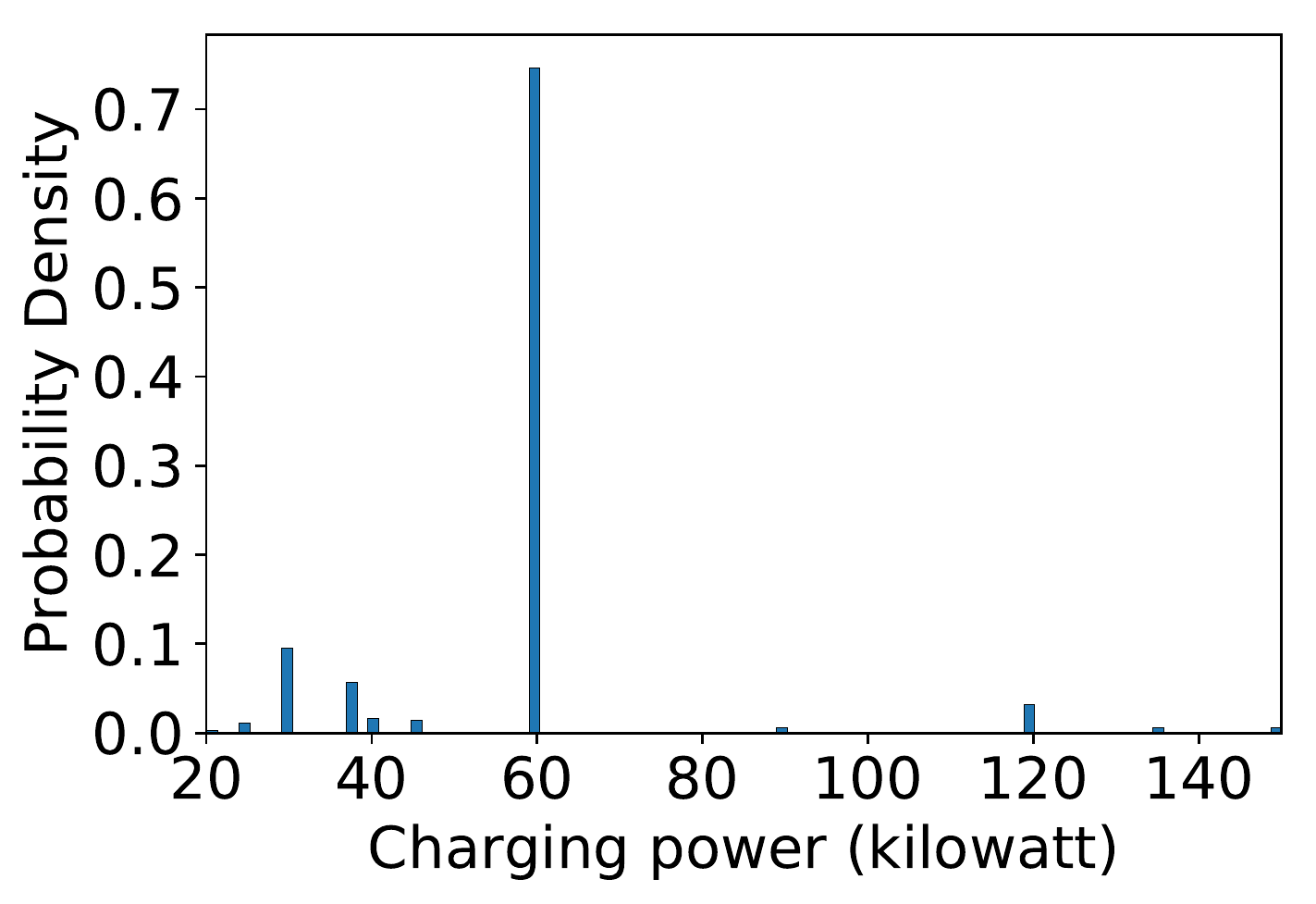}}
\subfigure[{\small CP}]{
	\includegraphics[width=0.48\columnwidth]{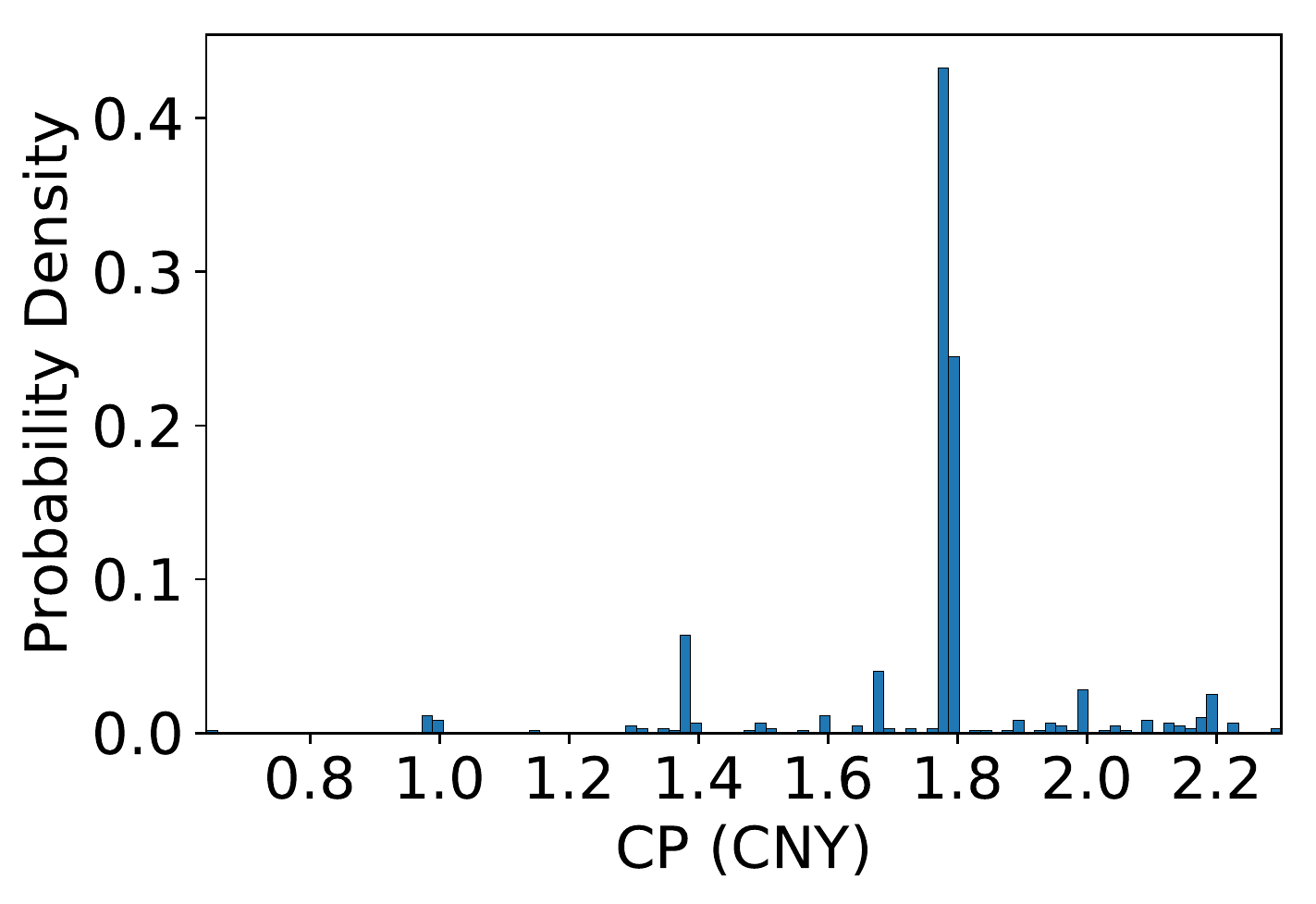}}
\caption{Distributions of charging requests quantity, ETA, charging powers, and CP on \beijing.}
\label{exp:data_dist}
\end{figure} 

\section{Dataset Statistical Analysis}
\label{app:data_analysis}
We take \beijing as an example for more statistical analysis. 
\figref{exp:data_dist} shows the distribution
of the quantities of charging requests in different time, and the distributions of ETA, charging powers, and CP. We can observe the charging requests quantity changes notably along with time in a day, which indicates unbalanced charging demands. Besides, we observe it is significantly different by comparing the distributions of ETA with CP, where the latter is much more concentrated. This observation indicates the large differences between the two objectives.

\section{Proposition}
\label{app:prop}
\begin{proposition}
\label{pf:avg_reward}
Define $r^1$~(\eg $r^{cwt}$) and $r^2$~(\eg $r^{cp}$) as two different reward functions, then the following two policy gradients to update the policy $\bm{b}$ are equivalent:
\begin{align}
\nabla_{\theta_{\bm{b}}}J_1(\bm{b})=\mathbb{E}_{s_t \sim D}\left[\nabla_{\theta_{\bm{b}}}\bm{b}(s|\theta_{\bm{b}})|_{s=s_t} \nabla_a Q_{\bm{b}}(s,a)|_{s=s_t,a=\bm{b}(s_t)}\right]
\end{align}
\begin{align}
\nabla_{\theta_{\bm{b}}}J_2(\bm{b})&=\mathbb{E}_{s_t \sim D}\left[\nabla_{\theta_{\bm{b}}}\bm{b}(s|\theta_{\bm{b}})|_{s=s_t} \nabla_a Q^1_{\bm{b}}(s,a)|_{s=s_t,a=\bm{b}(s_t)}\right.\\
&\nonumber\quad \left.+\nabla_{\theta_{\bm{b}}}\bm{b}(s|\theta_{\bm{b}})|_{s=s_t} \nabla_a Q^2_{\bm{b}}(s,a)|_{s=s_t,a=\bm{b}(s_t)}\right]\\
&\nonumber 
=\mathbb{E}_{s_t \sim D}\left[\nabla_{\theta_{\bm{b}}}\bm{b}(s|\theta_{\bm{b}})|_{s=s_t} \nabla_a \left(Q^1_{\bm{b}}(s,a)+Q^2_{\bm{b}}(s,a)\right)|_{s=s_t,a=\bm{b}(s_t)}\right]
\end{align}
\begin{align}
&\nonumber \text{where } Q_{\bm{b}}(s,a)=\mathbb{E}_{\bm{b}}\left[\sum_{i=t}^{\infty} \gamma^{(i-t)} \left(r^1(s_i,a_i)+r^2(s_i,a_i)\right) | s=s_t, a=a_t \right] \\
&\nonumber \text{and } Q^k_{\bm{b}}(s,a)=\mathbb{E}_{\bm{b}}\left[\sum_{i=t}^{\infty}\gamma^{(i-t)} r^k(s_i,a_i) | s=s_t, a=a_t\right], k=\{1,2\}.
\end{align}

\end{proposition}

\end{document}